%% file: Infus_ESS_revised.tex
\documentclass[final]{elsarticle}

\journal{}%
\usepackage[hmargin=3cm,vmargin=3cm]{geometry}

\usepackage{soul}
\usepackage{subfigure}
\usepackage{tikz}
\usepackage{lscape}
\usepackage{algorithm}
\usepackage{enumitem}\usepackage{float}
\usepackage{algpseudocode}
\usepackage{graphicx}
\usepackage{varioref}
\usepackage{array}
\usepackage{amsmath}
\usepackage{booktabs}
\usepackage{multirow}
\usepackage{array,arydshln}
\usepackage{amssymb}
\usepackage{hhline}
\usepackage{times,rotating}
\usepackage[utf8]{inputenc}

\usepackage{tabularx,ragged2e}
\newcolumntype{T}[1]{>{\raggedright\arraybackslash}p{#1}}
\newcolumntype{M}[1]{>{\centering\arraybackslash}m{#1}}

\newcolumntype{L}[1]{>{\raggedright\let\newline\\\arraybackslash\hspace{0pt}}m{#1}}
\newcolumntype{C}[1]{>{\centering\let\newline\\\arraybackslash\hspace{0pt}}m{#1}}
\newcolumntype{R}[1]{>{\raggedleft\let\newline\\\arraybackslash\hspace{0pt}}m{#1}} 

\usepackage{tikz}
\usepackage{forest}
\usetikzlibrary{positioning}

\usepackage{framed} 
\usepackage{multicol} 
\usepackage{nomencl} 
\makenomenclature
\renewcommand*\nompreamble{\begin{multicols}{2}}
\renewcommand*\nompostamble{\end{multicols}}

\newcommand{\red}[1]{\textcolor[rgb]{0.8,0,0}{#1}}

\usepackage{hyperref}
\hypersetup{
    bookmarks=true,         
    unicode=false,          
    pdftoolbar=true,        
    pdfmenubar=true,        
    pdffitwindow=false,     
    pdfstartview={FitH},    
    pdftitle={},         
    pdfauthor={}, 
    pdfsubject={},       
    pdfcreator={},  
    pdfproducer={}, 
    pdfkeywords={keyword1} {key2} {key3}, 
    pdfnewwindow=true,      
    colorlinks=true,        
    linkcolor=gray,         
    citecolor=gray,         
    filecolor=blue,         
    urlcolor=blue           
}

\begin{document}

\begin{frontmatter}
 
\title{Machine Learning Information Fusion in Earth Observation: A Comprehensive Review of Methods, Applications and Data Sources} 

\author[a]{S. Salcedo-Sanz\corref{cor1}}
\author[b]{P. Ghamisi}
\author[c]{M. Piles}
\author[d]{M. Werner}
\author[a]{L. Cuadra}
\author[c]{Á. Moreno-Martínez}
\author[g]{E. Izquierdo-Verdiguier}
\author[c]{J. Mu\~noz-Marí}
\author[e,f]{A. Mosavi}
\author[c]{G. Camps-Valls}

\address[a]{Universidad de Alcal\'a, 28871 Alcal\'a de Henares, Spain.}
\address[b]{Helmholtz-Zentrum Dresden-Rossendorf, Helmholtz Institute Freiberg for Resource Technology, Germany.}
\address[c]{Universitat de Val\`encia, 46980, Val\`encia, Spain.}
\address[d]{Technical University of Munich, Germany.}
\address[e] {Kando Kalman Faculty of Electrical Engineering, Obuda University, 1034 Budapest, Hungary.} 
\address[f] {School of the Built Environment, Oxford Brookes University, Oxford OX30BP, UK}
\address[g] {University of Natural Resources and Life Science (BOKU), 1190, Vienna, Austria}

\cortext[cor1]{Corresponding author: Sancho Salcedo-Sanz, Universidad de Alcal\'a, 28871 Alcal\'a de Henares, Spain. Phone: +34 91 885 67 31. sancho.salcedo@uah.es {\bf Preprint, paper published in Information Fusion. Published Volume 63, November 2020, Pages 256-272. https://doi.org/10.1016/j.inffus.2020.07.004}}

\begin{abstract}
This paper reviews the most important information fusion data-driven algorithms based on Machine Learning (ML) techniques for problems in Earth observation. Nowadays we observe and model the Earth with a wealth of observations, from a plethora of different sensors, measuring states, fluxes, processes and variables, at unprecedented spatial and temporal resolutions. Earth observation is well equipped with remote sensing systems, mounted on  satellites and airborne platforms, but it also involves {\em in-situ} observations, numerical models and social media data streams, among other data sources. Data-driven approaches, and ML techniques in particular, are the natural choice to extract significant information from this data deluge. This paper produces a thorough review of the latest work on information fusion for Earth observation, with a practical intention, not only focusing on describing the most relevant previous works in the field, but also the most important Earth observation applications where ML information fusion has obtained significant results. We also review some of the most currently used data sets, models and sources for Earth observation problems, describing their importance and how to obtain the data whether needed. Finally, we illustrate the application of ML data fusion with a representative set of case studies, as well as we discuss and outlook the near future of the field.
\end{abstract}

\end{frontmatter}

\clearpage
\tableofcontents
\clearpage

\section{Introduction}

The Earth is a highly complex, dynamic, and networked system where very different physical, chemical and biological processes interact, to form the world we know \cite{Kump04,Stanley05,Brasseur05}. The description of such a complex system needs of the integration of different disciplines such as Physics, Chemistry, Mathematics and other applied sciences, leading to what has been coined as Earth System Science (ESS) \cite{Jacobson00}. The analysis of the Earth system involves studying interacting processes occurring in several spheres (atmosphere, hydrosphere, cryosphere, geosphere, pedosphere, biosphere, and magnetosphere) as well as the anthroposphere. Earth system science provides the physical basis of the world we live in, with the final objective of obtaining a sustainable development of our society\footnote{\href{https://www.un.org/sustainabledevelopment/sustainable-development-goals/}{https://www.un.org/sustainabledevelopment/sustainable-development-goals/}}.

Traditionally, Earth system models characterize processes and their relations by encoding the known physical knowledge. This involves deriving models from first principles, which typically involves physics-based mechanistic modeling. Such Earth system models are complex constructs, and can be designed at different scales, but provide and encode the fundamental basis for understanding, forecasting, and modeling.

Models are often confronted with observations for refinement and improvement. In the last five decades, the field of Earth Observation (EO) from space has allowed monitoring and modelling the processes on the Earth surface, and their interaction with the atmosphere, by obtaining quantitative measurements and estimations of geo-bio-physical variables, and has permitted to detect extremes, changes, and anomalies. By combining EO data from stations, sensors, and ancillary model simulations, we can now monitor our planet with unprecedented accuracy, spatially explicitly and temporally resolved.

In the last decade, however, we have witnessed two important changes: the big data and the Machine Learning (ML) revolutions, which have impacted many areas of Science and Engineering, but also the Earth sciences \cite{Reichstein19}. Both combined are leading to paradigm shifts in the way we now study the Earth system:
\begin{itemize}
    \item {\em The big data revolution.} Nowadays, we observe and model the Earth with a wealth of observations and data measured from a plethora of different sensors measuring states and physical variables at unprecedented spatial, spectral, and temporal resolutions. They include remote sensing systems, mounted on satellites, airplanes, and drones, but also {\em in-situ} observations (increasingly from autonomous sensors) at, and below, the surface and in the atmosphere. Data from numerical, radiative transfer and climate models, and from reanalysis, are also very effective for tackling specific problems in Earth science. The data deluge is ever-growing in volume (hundreds of petabytes already), speed (estimated around 5 Pb/yr), variety (in sampling frequencies, spectral ranges, spatio-temporal scales and dimensionality), and uncertainty (from observational errors to conceptual inconsistencies).
    \item {\em The Machine Learning revolution.} Besides, ML techniques have emerged as a fundamentally effective way to model and extract patterns out of big data in a (semi)automatic manner. Earth science has been also impacted by such revolution in many different ways. For example, machine learning models are now routinely used to predict and understand components of the Earth system: 1) classification of land cover types, 2) modelling of land-atmosphere and ocean-atmosphere exchange of greenhouse gases, 3) detection of anomalies and extreme events, and 4) causal discovery have greatly benefited from ML approaches. ML has been also used to complement physical models developed in the last 50 years, which are now able to assimilate the measured data and yield more accurate estimates/predictions of the evolution of the system (or parts thereof), such as the Atmosphere or the Ocean. 
    Hybrid modeling and physics-aware machine learning are nowadays emerging fields too, and promise data-driven and physically consistent models for the future study of the Earth system.
\end{itemize}
All in all, the access to such unprecedented big data sources, increased computational power, and the recent advances in ML offer exciting new opportunities for expanding our knowledge about the Earth system from data. EO is now at the center of the data processing pipeline. Either data assimilation, canonical ML, or advanced hybrid modeling needs to successfully exploit at maximum the diversity and complementarity of the different data sources. The use of Information Fusion is thus essential to obtain robust models in this discipline, with physical coherence and high accuracy, able to describe the different processes the Earth system.

In any information fusion process involving EO data, heterogeneous data coming from very different sources are considered. It is thus understood that we are able to extract descriptors (features, covariates) that describe them, following a specific problem-dependent procedure, so these features can be further processed by applying algorithms. Depending on the problem tackled, these data features can present different spatio-temporal resolution, high dimension, or any other characteristic that makes their direct processing using algorithms (ML in this case) difficult. The different processes and methods for data fusion in EO can be broadly classified depending on the level where the fusion is done: at a sub-feature (data transformations to harmonize sources), at the feature level (direct fusion of data sets (or their resulted features) into the ML algorithms), or at the decision level (different processing paths with fusion at the decision level).

In this paper, we review the current state of information fusion data-driven algorithms based on ML techniques, for problems in EO data analysis. The proposed review and perspective paper has a clear practical intention: we first describe the most relevant previous works in the field, structuring the description into different applications of EO. We also describe the most relevant and important data sets and sources for EO problems, i.e., what are the most important data sources currently available for EO studies, and how a researcher can obtain them. Finally, a carefully selected set of case studies will complete the work by illustrating in a practical way different aspects of ML information fusion for EO problems. We conclude the paper with a general outlook and discussion about the near future of this particular field.

Following the objective of this paper, we have structured this work into three large blocks after this introduction, and a final section of conclusions, discussion and outlook. First, we provide a complete and updated literature review of ML information fusion in EO, including previous reviews and overview works, a taxonomy of the field and a review of illustrative previous works related to ML for information fusion. The second block of the paper is devoted to describe the existing data sources useful for EO, including satellite data, in-situ observations and different models applied to EO problems. The third block of the paper presents different case studies focused on ML information fusion in real EO problems. The final part of the paper consists of a discussion, conclusion, and perspective section, where the current challenges on the field, recent trends and possible future research are outlined.

\section{Machine Learning information fusion in Earth observation: a comprehensive literature review}

Earth observation is a huge research area, involving extremely different problems, applications and cases in all the spheres of the Earth system. The challenge of summarizing all the work that have been done in the whole area is therefore unmanageable. We, however, focus on ML information fusion for EO, which alleviates somehow the difficulty, though it is still very hard to summarize all the work on this field, and there are different ways to tackle this task. We have decided to structure this section based on applications and problems faced with ML information fusion in EO problems. We have therefore based this discussion of previous works on existing applications and problems tackled with ML information fusion. First, just to show the huge previous work carried out in the whole EO area in the last years, we discuss some previous reviews and overview works in topics somehow related to EO, but covering more specific or partial aspects, not directly related to ML information fusion. We present afterwards a taxonomy of ML information fusion approaches, and finally we carry out then a complete description of previous works dealing with ML information fusion in EO problems, classified by different problems or application areas.

\subsection{Previous reviews and overviews in EO}

The importance of Earth modeling and the current interest in related applications have led to a huge amount of research work in the last years, some of them focused on information fusion techniques, in different areas of the topic. The interest is of a such magnitude that a considerably high number of reviews and overviews have been published, the large majority within the last 5 years. Table~\ref{table:summary} summarizes the body of literature in the intersection of ML, Earth sciences, and information fusion applications.

The first review article focused on information fusion for Earth observational data is \cite{Wald00}. This seminal review paper summarized the main concepts related to information fusion in a general, coarse grain, approach. Another early review on topics involving Earth sciences is the one in \cite{Cherkassky06}, mainly focused on a broad description of computational intelligence methods applied to this field, with special emphasis on the importance of data fusion. The subsequent review articles are much more specific, focused on detailed parts of ESS or on data source/methods. Many of these reviews deal with different aspects of remote sensing. Regarding this, \cite{Torabzadeh14} has presented a review of fusion spectroscopy images and laser systems for forest ecosystem characterization, while \cite{Chen15} has recently analyzed the state-of-the art of spatio-temporal fusion models for remote sensing, by means of a comparison among the most important existing ones. Also within the framework of remote sensing, \cite{Gomez15} has presented a review on multi-modal classification techniques for remote sensing images, and \cite{Schmitt16} presented a review specifically focused on data fusion techniques and algorithms for remote sensing.

Image fusion methods and algorithms have been fully described in several recent review papers too:  \cite{Ghassemian16} presented a general review of image fusion for remote sensing, and \cite{Garzelli16} carried out a review on image fusion from the super-resolution paradigm. In \cite{Yokoya17,PG22}, the main methods and algorithms for hyperspectral and multispectral data fusion in remote sensing have been reviewed, whereas in \cite{Gham19} the basis, state-of-the-art and challenges of multisource and multitemporal data fusion algorithms are discussed.

Two recent review papers deal with Big Data methods in Earth sciences: \cite{Guo16} is focused on big data techniques from satellite data sources, whereas \cite{Gibert18} introduces environmental data science algorithms and methods, in a wide number of ESS applications. Finally, some reviews on deep learning with focus on Earth sciences have been recently published, such as \cite{Ball17,Ma19,PLi2019,Yuan20}, which describe the most recent deep learning approaches in remote sensing applications, or \cite{Zhang18} with a broader perspective in the big data area. Finally, it is worth mentioning the perspective paper \cite{Reichstein19} which, focused on describing the most important challenges and future directions in data-driven ESS, suggest that multisource information fusion and hybrid modeling will play a fundamental role in the near future.

\subsection{A taxonomy of ML information fusion approaches}

A variety of information fusion schemes have been proposed in the context of EO. Broadly speaking, information fusion is concerned on the multisource data combination and support decision making. Each fusion method is designed for a specific problem so it is challenging, if not impossible, to define a full taxonomical overview of the field. The main building blocks, however, have to do with the exploitation of i) disparate inputs, ii) data (pre)processing approach, iii) fusion mechanism, and iv) outputs post-processing. Actually, fusion approaches are usually named depending on the type of modalities, so a simple taxonomy of fusion problems can be defined in terms of {\em when, at what level, and how} the fusion is done. This is why we distinguish between the following types of fusion approaches:
\begin{enumerate}
    \item {\em sub-feature level}, which usually involve different spatial-temporal scales fused following appropriate transforms of the data with the aim to harmonize sources into a common ``multidimensional grids'';
    \item {\em feature level}, where a direct fusion of the data sets is simply stacked and fed into the ML of choice. This direct stacking can be more sophisticated if optimal feature combinations, and data source transformations, are learned from data to end up stacking feature representations;
    \item {\em decision level}, where one performs different processing paths for each modality, followed by fusion at the decision level. This assumes that the outputs can be combined to improve the achieved accuracy. Different methods exist here that optimally operate on the combination of output activation functions.
\end{enumerate}
The best way to understand such differences is however to present different real examples in the literature, as follows.

\subsection{Literature review}

In this section we review the most important previous works on ML information fusion in EO problems. In terms of ML algorithms, the field mainly exploits either {\em classifiers} (of land use or land cover), or {\em anomaly, target and change detection algorithms} (for screening or identifying one class of interest and discard the rest), or {\em regression methods} (to estimate a particular variable of interest from either sensory data mounted on satellites, airborne or drones). We have structured this section in different subsections by application area or problem type, taken into account the most usual problems in which ML information fusion techniques have obtained significant results.

\subsubsection{Surface temperature}
The accurate estimation of surface temperature (inland and sea) from different sensors, both grounded and satellites, is extremely important in a number of problems, including agricultural studies, energy balance, land desertification and climate change applications, among others. In the literature, it is possible to find different studies applying data driven and ML techniques together with information fusion methods to estimate surface temperature. In \cite{Ortiz12} a feature level information fusion algorithm which hybridizes local atmospheric variables information from a ground station with synoptic information from numerical models is proposed for temperature prediction at Barcelona airport, Spain. An ensemble of Support Vector Regression (SVR) algorithms is used to carry out the information fusion and to obtain the temperature prediction. In \cite{Moosavi15}, a hybrid wavelet ML feature level fusion approach is proposed to obtain high-resolution land surface temperature, mixing Landsat 8 thermal bands and MODIS (moderate-resolution imaging spectroradiometer) pixels. Wavelets Support Vector Regression, adaptive network-based fuzzy inference system (ANFIS) and neural networks are the artificial intelligence methods tested in this problem. In \cite{Xia19}, a problem of high-spatiotemporal-resolution land surface temperature reconstruction is tackled, by applying a weighted combination kernel-based and fusion methods, in order to improve the spatial and temporal resolutions of satellite images. This method can be classified as a sub-feature level approach. Specifically, MODIS and Landsat 8 datasets have been considered for the experimental evaluation of the proposed method, obtaining more accurate images than the kernel method on its own. Recently, in \cite{Zhang20} a feature level information fusion process from different sources (measuring points) and a gated neural network was proposed to estimate the sea temperature at Bohai Sea, China.

\subsubsection{Droughts and water quality}

Closely related with surface temperature estimation, the analysis of drought and water quality using ML and data fusion techniques has been recently proposed. In \cite{Park17}, a high resolution soil moisture drought index was proposed. This index is based on measurements of the Advanced Microwave Scanning Radiometer on the Earth Observing System over the Korea, improved by MODIS and Tropical Rainfall Measuring Mission satellite sensors information, which was used to carry out a high-resolution feature level downscaling with a Random Forest (RF) algorithm to 1 Km measurements. In \cite{Yao17} a SVR algorithm was proposed to obtain an accurate estimation of Evapotranspiration by fusion of three process-based Evapotranspiration algorithms: MOD16, PT-JPL and SEMI-PM, to produce a feature level information fusion approach.

In \cite{Alizadeh18}, the study of drought events was carried out by applying new fusion approaches from high-resolution satellite and reanalysis data at feature level. The work in \cite{Feng19} is focused on determining whether the fusion of several remotely-sensed drought factors could be effectively used for monitoring drought events in Australia. This problem is tackled as a regression task with information fusion at feature level, where three ML approaches have been tested, RF, SVR and artificial neural networks.

Regarding ML fusion methods for water quality studies, in \cite{Dona15} a Genetic Programming approach is applied to fusion data from different satellite sources, such as MODIS or Landsat Thematic Mapper. The objective is to generate daily estimates of different water quality parameters such as chlorophyll-{\em a} concentrations or water transparency, for a freshwater lake (Albufera) in Valencia, Spain. Also, \cite{Jiang18} proposes a general framework based on computer vision feature representation, for the fusion of multisource spatiotemporal data for hydrological modeling (sub-feature level information fusion), following by the application of artificial neural networks and SVR algorithms. Finally, a review of fusion methods in water quality can be found in \cite{Sagan20}.

\subsubsection{Cloud detection and classification}
Cloud detection and classification is an EO area where ML information fusion has been successfully applied. In \cite{Liu18} a Convolutional Neural Network (CNN) has been applied to a problem of cloud classification using ground-based images. This proposal uses a two-stream structure which contains the vision subnetwork and multimodal subnetwork. In another layer of the network (fusion layer), the visual and multimodal features from the two subnetworks are extracted and then integrated using a weighted strategy. This is therefore a decision level method. The results have been tested in a specific database of multimodal ground-based clouds. In \cite{Li19} a cloud detection method based on CNN was proposed. The idea is to extract multi-scale and high-level spatial features by using a symmetric encoder-decoder module. The feature maps of multiple scales are then passed to a multi-scale feature fusion module, designed to fuse the features of different scales for the output in another decision level process. A final binary classifier is able to obtain the final cloud and cloud shadow mask. The method was validated on a large number of optical satellite images around the world, with different spatial resolutions ranging from 0.5 to 50 m.

\subsubsection{Land use applications}

Land use is another important area of interest in EO where information fusion has emerged lately. Recently, there have been many different works on land use tasks which combine different data sources or fusion algorithms with ML, in order to obtain robust approaches with optimal performance. For example, \cite{Wang16} proposes the fusion of multispectral HJ1B imagery (from China's HJ-CCD B satellite) and ALOS (Advanced Land Observing Satellite) PALSAR L-band (Phased Array type L-band Synthetic Aperture Radar) data for land cover classification, using sub-feature level information fusion, Support Vector Machine (SVM) and RF algorithms.
In \cite{Puttina17} an investigation focused on extraction of buildings from middle and high resolution satellite images is carried out, by fusing the information of different spectral indices with ML algorithms and sub-feature level information fusion techniques.

In \cite{Guy18} a deep CNN paradigm has been applied to a problem of automatic-land use classification from satellite images. Besides, in \cite{Lu19} a cellular automata -- Markov model is proposed in order to generate land use images from fusing images belonging to different years. This approach considers a sub-feature level information fusion mechanism, and has been successfully tested in reconstructing land use images in Hefei (China), from satellite data from the last 30 years. In \cite{Rasaei19} a feature level approach for fusing soil data coming from legacy soil surveys with direct soil information from remote sensing images was proposed. In \cite{P197}, several decision level and feature level fusion approaches were developed to tackle the problem of local climate zones classification based on a multitemporal and multimodal dataset, including image (Landsat 8 and Sentinel-2) and vector data (from OpenStreetMap) using ensemble classifiers and deep learning approaches. Finally, in \cite{Shaharum20} a feature level fusion of information from Google Earth and ML algorithms, including SVMs and regression trees, was proposed for a problem of palm oil mapping in Malaysia.

\subsubsection{Image classification and segmentation}

Image classification is another field closely related to very different remote sensing applications, where ML information fusion has been successfully applied. In \cite{Alajlan12} a decision level method which combines SVM and fuzzy C-means clustering for fusing hyperspectral images information is proposed. In this approach, the SVM is used to generate a spectral-based classification map, whereas the fuzzy C-means is used to provide an ensemble of clustering maps. In  \cite{Ghamisi17} a feature level fusion of hyperspectral and light detection and ranging (LiDAR) images is proposed, under the hypothesis that LiDAR provide a source of complementary information, which can be really useful to improve classification of hyperspectral data. Specifically, the derived features from the two sources are fused via either feature stacking or graph-based feature fusion, and the fused features are fed to a deep CNN with logistic regression to produce the final classification map. The fusion of hyperspectral data and LiDAR is also treated in \cite{Khoda15}, and mixing hyperspectral data, LiDAR and other data sources with CNN in \cite{Xu18}. Finally, in \cite{Liang18} an unsupervised feature extraction method based on deep multiscale spectral-spatial feature fusion for hyperspectral images classification is proposed. The method is based on pre-trained filter banks and on a new unsupervised cooperative sparse autoencoder method to fuse together the deep spatial feature and the raw spectral information (sub-feature level fusion).

\subsubsection{Renewable Energy}

Renewable energy resources are fully related to ESS, since the main renewable sources (wind, solar, ocean) are fully conditioned by atmospheric or oceanic conditions. Due to their intermittent intrinsic nature, renewable energy sources present difficulties to be integrated in the energy mix, and usually need prediction techniques to this end. Interestingly, the main renewable energies are affected by climate change, which produces a redistribution of the renewable resources. In general, the study of renewable resource prediction is a hot topic in which the fusion of different sources of information has also been explored. In fact, information fusion in renewable energy has been mainly exploded in solar energy prediction systems. Solar energy resource prediction is fully connected to the prediction of clouds, which is a difficult problem from the meteorological point of view, in which information from different data sources is able to improve the prediction systems. 

In \cite{Mellit08}, several neural network architectures with different input data sources were proposed to problems of solar radiation. More recently, in \cite{Salcedo14} a solar radiation prediction problem from different sources data was proposed. Data from in-situ measurement and from the GFS model were the inputs for a temporal Gaussian Process algorithm, which obtained better results than alternative ML algorithms in the problem. Another work dealing with data from different sources in a problem of solar radiation prediction is \cite{Mazorra16}, which proposed the prediction of intra-day solar radiation by means of data from irradiance measurements, satellite data and weather prediction models using artificial neural networks. Results in two different points of the Canary islands were reported. In \cite{Urraca20} an study on the evaluation of global horizontal irradiance from two different Reanalysis projects (ERA5 and COSMO-REA6) is carried out. This work uses ground and satellite-based data to estimate the accuracy of these reanalysis in obtaining horizontal irradiance. The results obtained show that Reanalysis data have important absolute error when estimating the irradiation, due to an inefficient cloud cover estimation. However, reanalysis products are an important data source for prediction and estimation problems in renewable energy, useful for information fusion with other sources, as in \cite{Babar20}, where data from ERA5 reanalysis are hybridized with satellite measurements at feature level, and all the information is processed with a RF algorithm in order to obtain an accurate estimation of solar irradiance at high latitudes.

\subsubsection{Model-data integration and assimilation}

Any optical remote sensing data present significant amounts of  noise and  missing data due to clouds, cloud shadows, and aerosol contamination, which difficult its use in any subsequent application. In addition, optical remote sensing sensors are hampered by a limited temporal, spectral or spatial resolutions \cite{thenkabail2018advanced}. As an example,  medium spatial resolution sensors, such as Landsat or the Sentinel 2, have low temporal resolutions (16 and 8 day revisit cycle respectively)  causing missing values continue to be one of the major limitations for their operational use, especially in areas with moderate to high cloud occurrence. The mitigation of undesired inherent data noise and minimizing the amount of missing data present are mandatory tasks in almost any application, since they are incompatible with a robust remote monitoring framework of earth's surface.

Because of the importance of this topic, the available scientific literature is rich in methods to deal with these issues, and solutions vary significantly with the different levels of sophistication \citep{kandasamy2013comparison}. Temporal, spatial, spatio-temporal, and blending (sensor fusion) approaches have been very valuable tools to reduce noise and recover missing pixels information, being data fusion methods very interesting approaches for overcoming individual sensor's limitations and combining different multiresolution datasets.

Most of data fusion approaches need to compute complex spatial operations to account for inhomogeneities within coarser spatial resolution pixels  \citep{gao2006blending}. These operations are computationally demanding allowing only the application of these algorithms to small areas. In the context of modern data assimilation approaches, Sedano \textit{et al.} \citep{sedano2014Kalman} introduced a pixel-based method with a Kalman filter (KF) \citep{Kalman1960new} to fuse time series of MODIS and Landsat vegetation indices. This KF implementation obtained satisfactory results but also allowed to account for realistic uncertainties in its calculations. Moreover,  the KF does not require explicit parameter tuning and it scales well in large scale applications due to its pixel-based nature.

\subsubsection{Unstructured domain data fusion}

The number of EO platforms for capturing remotely sensed data has been exponentially increased, ranging from an ever-growing number of satellites in orbit and planned for launch, to new platforms for capturing fine spatial resolution data such as unmanned aerial vehicles (UAVs). Moreover, a great interest has been recently dedicated to the new sources of ancillary data, to name a few, social media, crowd sourcing, scraping the internet and so on \cite{P7,P9}. These data have a very different modality to remote sensing data, but may be related to the subject of interest and, consequently, may be found useful with respect to specific problems. For example, social media data can provide local and live/real-time information suitable for accurate monitoring of our living environment \cite{P192}, in particular in a variety of applications relevant to smart cities \cite{P193,P194}, emergency and environmental hazards \cite{P195,P196}, among others.

\subsubsection{Other applications}

There are other applications of ML algorithms in EO which exploit data fusion to improve their results. One of the first works on exploiting data fusion from multi-sensor sources with ML was \cite{Fisher98}, where a problem of ionograms inversion is tackled using data fusion techniques with neural networks. More recently, there have been alternative specific applications, such as \cite{Carro17} where a problem of total ozone in column prediction is tackled with SVMs and information fusion from different data sources, such as numerical models, ground stations and satellite data. In \cite{Du19} a problem of eddy detection from different satellite sensors images was faced, using a deep learning approach for multi-scale feature fusion, followed by a SVM algorithm.
In \cite{Kattenborn15} the combination of data from multiple sensors was used to improve the estimation of forest biomass. Data from interferometric and photogrammetric based predictors, in combination with hyperspectral predictors were used, by applying ML algorithms such as RF, Generalized Additive Models or boosted algorithms. These approaches were applied to estimate biomass in a temperate forest near Karlsruhe, Germany. In \cite{Effrosy18} different ML classifiers were applied to a problem of detecting seagrass presence/absence and distinguishing seagrass families in the Mediterranean, from fusion of seagrass presence data and other external environmental variables. As a final application of ML information fusion in ESS, in \cite{Li18b}, a problem of environmental event sound recognition is tackled by means of a staked CNN.

\begin{table}[!ht]
\scriptsize
\vspace{12pt}
\caption{Summary of Information Fusion works that use Machine Learning in Earth observation applications. For each group of applications (first column) we have summarized the main points concerning the type / level of information fusion (second column) and the different ML techniques used (on the third one). The last column groups the corresponding references for any set of applications.}

\centering
\begin{tabular}{| p{2.5cm} |   p{7.5cm} | p{4.0cm}| p{1cm}|}
\hline
 \textbf{Approach and / or Application}   &  \textbf{Type  of data fusion / Level of Information Fusion}  &  \textbf{Machine Learning algorithms}    & \textbf{Ref.} \\
\hline
General reviews on ESS &   Fundamentals of Information Fusion to properly process Earth data sets.  \cite{Wald00}. Information Fusion at several levels applied to climate, weather, geophysical and hydrologic problems \cite{Cherkassky06}. &   Neural, Fuzzy and Evolutionary Computation \cite{Cherkassky06} &  \cite{Wald00,Cherkassky06} \\
\hline

Reviews on applications of remote sensing  that focus on different data/level fusion.   &  Need for information fusion at different levels in deep learning approaches to remote sensing in EO applications  \cite{Ball17,Ma19,Zhang18}. Fundamentals and challenges of multisource and multitemporal data fusion algorithms \cite{Gham19,Guo16}. Comparison of the four most relevant spatio-temporal fusion models \cite{Chen15}. Review on different types of multi-modal image fusion for classification at subpixel level, pixel level, feature level, and decision level \cite{Gomez15}. Review on data fusion approaches for remote sensing \cite{Schmitt16}. Discussion of super-resolution solutions for spatio-temporal fusion and pan-sharpening \cite{Garzelli16}. Review on spectral and spatial information fusion for hyperspectral data classification \cite{PG22}. Analysis of hyperspectral and multispectral data fusion at several levels \cite{Yokoya17,PLi2019}. Hyperspectral fusion of images at decision level \cite{Alajlan12,Ghamisi17,Khoda15,Xu18}. Multiscale spectral-spatial feature fusion for hyperspectral images at sub-feature level fusion \cite{Liang18}. &   Neural Networks \cite{Gomez15,Ghassemian16,Ghamisi17}, Extreme Learning Machines \cite{Gham19}, Support Vector Machines  \cite{Gomez15,Ghassemian16,Yokoya17,Gham19}, Deep learning  \cite{Gomez15,Ghassemian16,Gham19,PLi2019,Yuan20,Ball17,Ma19,Zhang18,PG22},  Fuzzy C-means Clustering \cite{Alajlan12}, Convolutional Neural Networks  \cite{Ghamisi17,Xu18,PG22}, Unsupervised cooperative sparse auto-encoder method \cite{Xu18}.    & \cite{Ball17,Ma19,Zhang18,Gham19,Guo16,Chen15,Gomez15,Garzelli16,Yokoya17,PLi2019,Yuan20,Alajlan12,PG22,Ghamisi17,Xu18,Liang18} \\
\hline

 Combining deep learning and process-based approaches for Earth System Science  &      Multi-source, high-dimensional, multi-scale,  complex spatio-temporal, interrelated data \cite{Reichstein19}.   &  Discusses deep-learning challenges in ESS. Suggests that future models should integrate process-based and machine learning approaches  & \cite{Reichstein19} \\
 
 \hline
Surface Temperature  &  Information fusion is carried out at feature level \cite{Ortiz12,Moosavi15,Zhang20} and sub-feature level \cite{Xia19}. &  Support Vector Regression  \cite{Ortiz12}, ANFIS \cite{Moosavi15}, kernel methods \cite{Xia19}, Neural Networks \cite{Zhang20}. & \cite{Ortiz12,Moosavi15,Xia19,Zhang20}.   \\
\hline
Droughts events    &  Information fusion is carried out at feature level \cite{Park17,Yao17,Alizadeh18,Feng19}. &  Random Forest \cite{Park17}, Support Vector Regression, Neural Networks \cite{Yao17,Alizadeh18,Feng19}. & \cite{Park17,Yao17,Alizadeh18,Feng19}.   \\
\hline
Water quality  &     Different fusion data from satellites \cite{Dona15}, sub-feature level information fusion \cite{Jiang18}. &  Genetic Programming \cite{Dona15}, Neural Networks and Support Vector Regression \cite{Jiang18}, review of recent techniques \cite{Sagan20}. & \cite{Dona15,Jiang18,Sagan20}   \\
\hline
Cloud detection and classification    &   Image fusion at several levels.  &  Convolutional Neural Networks \cite{Liu18,Li19}  & \cite{Liu18,Li19}.   \\
\hline
Land use &    Multispectral satellite images, at sub-feature level information fusion \cite{Wang16,Puttina17,Guy18,Lu19}.  Feature level approach for fusing soil images \cite{Rasaei19}. Feature level approach for palm oil mapping \cite{Shaharum20}. &  Support Vector Machines \cite{Wang16,Shaharum20}, Random Forests algorithms \cite{Wang16}, Regression trees \cite{Shaharum20}, Deep Convolutional Neural Networks \cite{Guy18} in  a cellular automata -- Markov  \cite{Lu19}. & \cite{Wang16,Puttina17,Guy18,Lu19,Rasaei19,Shaharum20}.   \\
\hline
Forest parameters estimation. & Combination of airborne laser scanning and imaging spectroscopy data at data level \cite{Torabzadeh14}. Fusion of seagrass presence data and other external environmental variables \cite{Effrosy18}.  &  Support Vector Machines \cite{Torabzadeh14}. Random Forests, Generalized Additive Models or boosted algorithms \cite{Kattenborn15}. & \cite{Torabzadeh14,Kattenborn15,Effrosy18}. \\
\hline
Local climate zones classification &      Several decision level and feature level fusion approaches,  based on a multitemporal and multimodal images. &  Ensemble classifiers and deep learning approaches & \cite{P197}.   \\
\hline

Solar radiation prediction in renewable energy &  Irradiance measurements, reanalysis products and satellite data \cite{Mellit08,Salcedo14,Mazorra16,Babar20}.    &  Neural Networks \cite{Mellit08,Mazorra16}, Gaussian Process algorithm \cite{Salcedo14}, Random Forest \cite{Babar20}.  &  \cite{Mellit08,Salcedo14,Mazorra16,Babar20}  \\
\hline

Model-data integration and assimilation  &    Pixel-based method to fuse time series of MODIS and Landsat vegetation indices.  &  Kalman filter (KF)  &  \cite{sedano2014Kalman}  \\
\hline
Other applications  &  Ionograms inversion \cite{Fisher98}, Total ozone atmospheric content \cite{Carro17}, Eddies detection at sea \cite{Du19},  Forest Biomass estimation \cite{Kattenborn15}, Seagrass presence \cite{Effrosy18}, Environmental sound recognition \cite{Li18b}  & Neural networks \cite{Fisher98}, Support Vector Machines \cite{Carro17,Du19}, Random Forest \cite{Kattenborn15}, Convolutional Neural Networks \cite{Li18b}  &  \cite{Fisher98,Carro17,Du19,Kattenborn15,Effrosy18,Li18b}  \\
\hline
\end{tabular}
\label{table:summary}
\vspace{24pt}
\end{table}


\section{Data sources and models for Earth observation}\label{Data_sources}

The study of the ESS and its many components is based on observational data from very different sources, and very different atmospheric and climate models and simulations, among other data sources (such as social media or socio-economic data, sometimes fused with them). Every minute, millions of sensors collect data from the whole planet. Their processing, fusion with physic-based models, analysis and study is in the core of many works on EO-related problems. In this section, we describe the most important data sources and models currently available for studies on EO. We also provide the complete reference to the majority of these data sources, so the interested reader knows where they can be obtained. We have structured the section into subsections describing satellite, in-situ (ground-based, atmosphere and marine observations sources), forecasting models and finally reanalysis projects. 

\subsection{Satellite observations}

\begin{figure}[h!]
\centerline{\includegraphics[height=4.1cm]{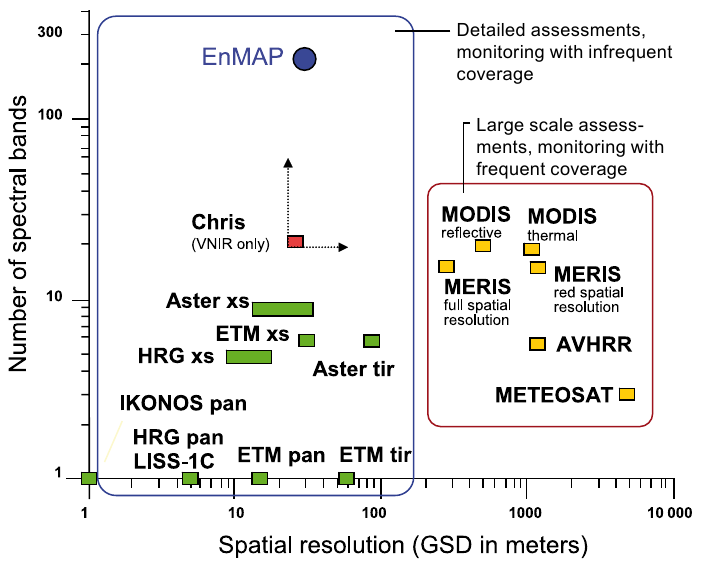} \hspace{0.5cm}
\includegraphics[height=4.1cm]{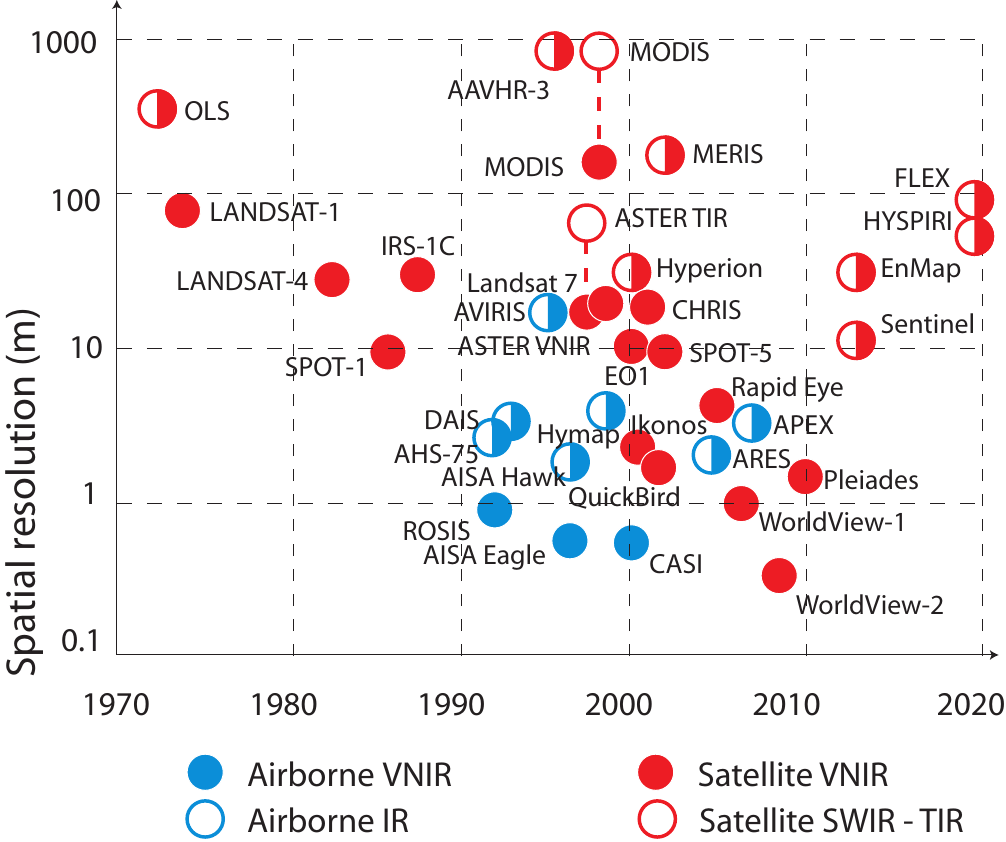}
\hspace{0.5cm}
\includegraphics[height=4.1cm]{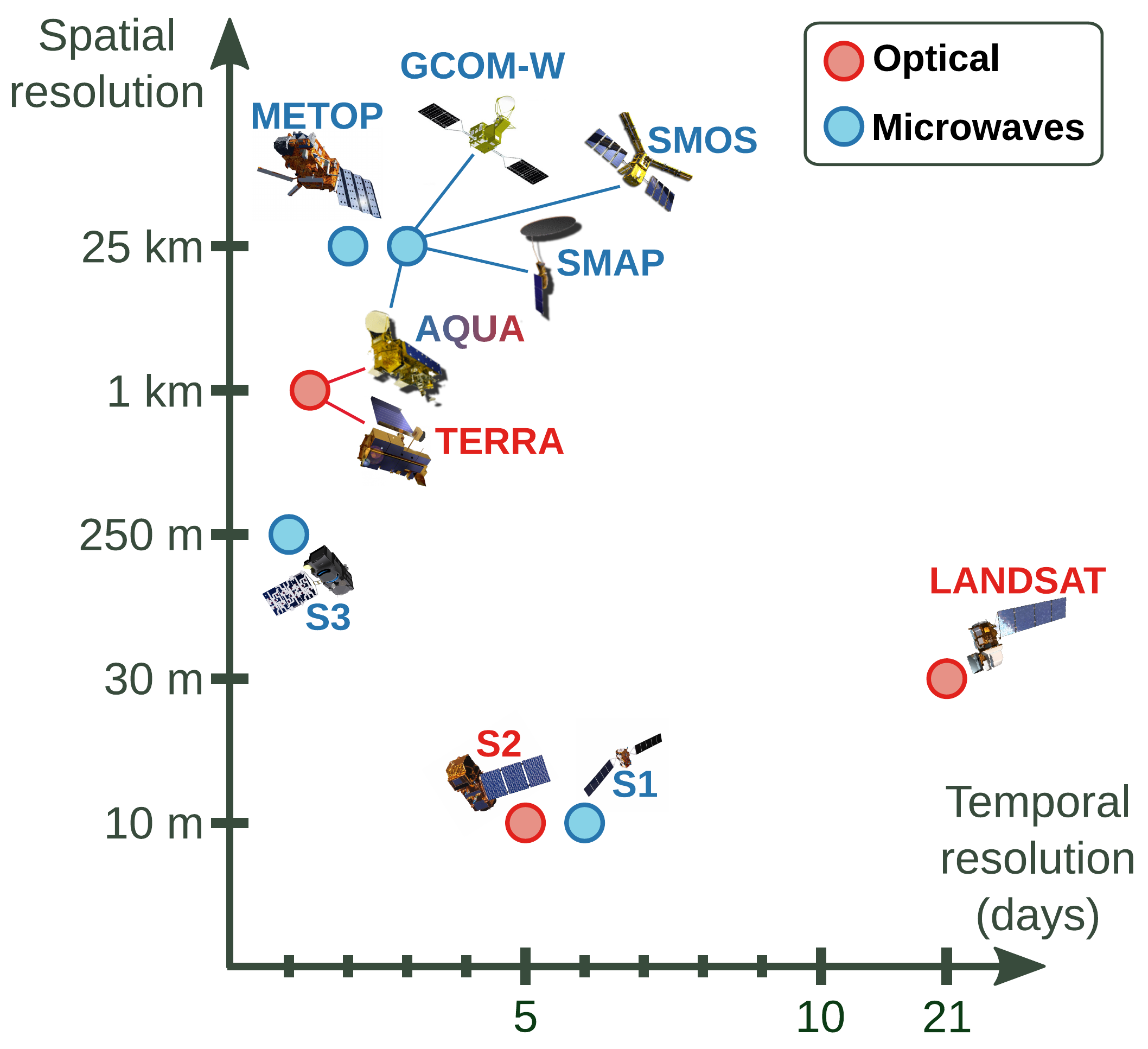}}
\vspace{-0.25cm}
\caption{{\em Left:} Performance comparison of the main air- and space-borne multi- and
hyperspectral systems in terms of spectral and spatial resolution.
{\em Middle:} Evolution of the spatial-spectral resolution through the years.
{\em Right:} Spatial and temporal coverage for some optical and microwaves satellite sensors.
Credits: http://www.enmap.de/.
}
\label{fig:sat_comparison}
\end{figure}

Attending to the type of {\em energy sources} involved in the data acquisition, two main kinds of remote sensing imaging instruments can be distinguished: either {\em passive} optical remote sensing, which relies on solar radiation as the illumination source \cite{Danson95,Richards99,Ustin04,Liang04,lillesand08}, or {\em active} sensors, where the energy is emitted by an antenna towards the Earth's surface and the energy scattered back to the satellite is measured \cite{Mott07,Wang08}. Some examples of passive sensors are infrared, charge-coupled  devices, radiometers, passive microwave, and multi and {hyperspectral} sensors \cite{shaw02}. On the other hand, in {Radar} systems, such as {Real Aperture RAR (RAR)} or {Synthetic Aperture Radar (SAR)}, are examples of systems for active remote sensing. Figure~\ref{fig:sat_comparison} shows some characteristics (spatial, spectral, and temporal resolutions) of the main available optical and microwave satellite sensors.

\subsection{In-situ observations: ground, atmosphere, and ocean}

\subsubsection{Ground-based observations}\label{Ground-based}
 
Ground-based observations have been traditionally the most basic source of atmospheric data, especially before the satellite Era. Currently, ground-based observations and satellites are mainly assimilated by numerical and climatic models, so its importance is still very high in EO applications, mainly in meteorological and climatological studies.

\begin{itemize}
    \item The European Climate Assessment \& Dataset project (ECA\&D) dataset \cite{Klein02,ECAD}. The ECA\&D database consists of daily station series of different meteorological/climatological variables: daily mean temperature, precipitation, sea level pressure or wind speed, among others. A gridded version with daily temperature, precipitation and pressure fields is also available for this database.

    \item The Climatic Research Unit (CRU) temperature datasets \cite{CRU}. These databases, usually known as CRU and HadCRUT4, are global temperature datasets, providing gridded temperature anomalies across the world.

    \item The Global precipitation data from the Global Precipitation and Climatology Center (GPCC) \cite{GPCC}. The GPCC provides gridded quality controlled station data. Another source of precipitation data is the database from the Global Precipitation Climatology Project (GPCP) \cite{GPCP}, which provides monthly precipitation dataset from 1979 to the present, combining observations and satellite precipitation data into $2.5^{\circ} \times 2.5^{\circ}$ global grids.

    \item The Global Historical Climatology Network-Monthly (GHCN-M) database \cite{GHCN}. The CHCN-M provides gridded land precipitation and temperature anomalies on a $5^{\circ} \times 5^{\circ}$ basis for the entire globe, from 1900 to 2015. This database is useful for climate monitoring activities, including calculation of global land surface temperature anomalies and trends.

\item The Goddard Institute for Space Studies (GISS) database \cite{GISS}. The GISS surface temperature database provides a measure of the changing global surface temperatures, with monthly temporal resolution since 1880. The data is available on an $2.5^{\circ} \times 2.5^{\circ}$ and two smoothing levels of 250km and 1200km smoothing. A land only version is also available. The dataset is continuously updated, and it is necessary to take into account that there are missing data values within this database.
\end{itemize}
There are other gridded ground-based databases at NOAA Earth System Research Laboratory (ESRL), which can be explored and downloaded from \cite{ESRL}. When it comes to large data volumes, the U.S. Government's open data initiative  \href{http://www.data.gov}{DATA.GOV://www.data.gov} collects and harmonizes all kind of social, economical, environmental, remotely-sensed datasets.

\subsubsection{Observations and simulations of land-atmosphere interactions}

Monitoring the land-atmosphere interactions is currently done thanks to a global network of continuous measurements. This is called the \href{https://fluxnet.fluxdata.org/}{FLUXNET} which is a global network of micro-meteorological tower sites that use eddy covariance methods to measure the exchanges of carbon dioxide, water vapor, and energy between the biosphere and atmosphere. FLUXNET is a global ``network of regional networks'' that serves to provide an infrastructure to compile, archive and distribute data for the scientific community. The large-scale measurement network, FLUXNET integrates site observations of these fluxes globally and provides detailed time series of carbon and energy fluxes across biomes and climates \cite{Baldocchi2008}. The data have been used to perform local or regional studies, but also to upscale globally the fluxes: move from point-based flux estimates to spatially explicit gridded fields of carbon and energy fluxes with machine learning and information fusion techniques~\cite{Tramontana16bg,Jung18fluxcom}.

An interesting initiative on land and atmosphere data harmonization is the Earth System Data Lab (ESDL) platform, \href{http://earthsystemdatalab.net/}{ESDL}, which curates a big database with more than 40 variables to monitor the processes occurring in our Planet. They are grouped in three data streams (land surface, atmospheric forcings but also socio-economic data~\cite{Mahecha19esdc}) and allow running algorithms in the web platform.

\subsubsection{Atmospheric observations}\label{atmos-based}

Atmospheric soundings data \cite{RUC,UWYO} are useful instruments to obtain an instant state of the atmosphere, i.e. a vertical profile of the atmosphere at a single point in time and above a particular position on Earth. Usually, a small instrument package called {\em radiosonde} is embedded into to a weather balloon, which is released from the surface and usually reaches the troposphere. The radiosonde is able to measure different properties of the atmosphere such as the vertical profile of temperature, dew point, wind speed and direction, among others, as it ascends. Atmospheric soundings data are very useful tools for EO, mainly in meteorological problems, specially in those related to local phenomena prediction, such as convection initialization or cloud formation, etc.

\subsubsection{Marine  observations}\label{marine-based}

Marine-based data sources are also freely available in many cases, and contribute to the study of different parts of the ESS. One of the main sources of physical oceanographic/Meteorology data is the data base of the National Data Buoy Center of the USA (NDBC) \cite{NDBC}. This database contains freely available data from dozens of ocean buoys located at the Atlantic Ocean, Caribbean Sea, West Coast of the USA and Alaskan Gulf. There are other important data sources for marine observations, such as the Pan-European Infrastructure for Ocean \& Marine Data Management \cite{PEIOMDM} in Europe or Australian Ocean Data Network \cite{OpenAccAus} for the Asia-Pacific region.

Over the last decade, the development of cutting-edge robotic technology has dramatically demonstrated the potential of autonomous observations to overcome the issue of data scarcity. For example, the ground-breaking Argo international program has set up an array of more than 3500 profiling floats that provide measurements of temperature and salinity profiles from the surface down to 2000 m below sea level every 10 days. As the first-ever global in-situ ocean-observing network in the history of oceanography, \href{http://www.argo.ucsd.edu/}{Argo} provides a crucial complement to satellite systems, thus enabling observation, understanding and prediction of the ocean's functioning and its role in the Earth's climate.

\subsection{Numerical weather models}\label{Forecasting_models}

Weather prediction models, also known as numerical weather models, solve systems of differential equations (the Navier-Stokes equations, energy, mass and linear momentum conservation) to obtain the future state the atmosphere. Specifically, fluid motion, thermodynamics, radiative transfer, and atmospheric chemistry are taken into account, using a coordinate system which divides the whole planet into a grid. Thus, wind speed, heat transfer, solar radiation, relative humidity, among other variables are calculated within each grid node, and the interactions with neighboring nodes are then used to estimate the atmospheric evolution for the future.

A few global forecasting models are run in the world, using current weather observations relayed from radiosondes, weather satellites and other observing systems as inputs (data assimilation process). Processing the vast datasets and obtaining the solution of the system of differential equations previously mentioned in each node of the global grid require of powerful supercomputers. Even in this case, the forecast skill of current numerical weather models extends to only about six days, due to the non-linear and chaotic nature of the atmosphere. Some of the global forecasting models currently in operation are:
\begin{itemize}
\item The Global Forecast System (GFS) is produced by the National Centers for Environmental Prediction (NCEP) \cite{Kanamitsu89,Kanamitsu91}.

\item The Global Environmental Multiscale Model (GEM), often known as the CMC model in North America, is an integrated forecasting and data assimilation system developed in the Recherche en Pr\'evision Num\'erique (RPN), Meteorological Research Branch (MRB), and the Canadian Meteorological Centre (CMC) \cite{Cote98,Cote98b,Yeh02}.

\item The Navy Global Environmental Model (NAVGEM) is a global numerical weather prediction computer simulation run by the United States Navy's Fleet Numerical Meteorology and Oceanography Center \cite{Hogan14}.

\item The Integrated Forecasting System (IFS), is the global numerical model run by the European Center for Medium-range Weather Forecast (ECMWF) \cite{ECMWF}.

The Unified Model (UM) is a numerical model of the atmosphere developed by the Met. Office (UK) \cite{Walters17}. It can be used for both weather and climate applications.

The ARPEGE (Action de Recherche Petite Echelle Grande Echelle) model is the operational numerical weather model at M\'et\'eo France \cite{Deque94}. This system was developed and it is currently maintained in collaboration with the ECMWF.

\end{itemize}

There are some works which have evaluated and compared the performance of these and other weather numerical models in different EO application contexts \cite{Buizza05,Perez13}. Recently, the possibility of using ML algorithms as alternative techniques for global weather forecasting has also been discussed \cite{Dueben18}.

\subsubsection{Reanalysis projects}\label{sec:Reanalysis}

A {\em Reanalysis} project is a methodology that combines existing past observations, by applying data assimilation techniques, with modern numerical weather models. Reanalysis projects usually extend over several decades and cover the entire planet (global) or extended regions (regional), being a very useful tool for obtaining a comprehensive picture of the state of the Earth system, which can be used for meteorological and climatological studies. There are several reanalysis projects currently in operation, and some others which were the precursors of the current ones.

Two of the first reanalyses projects in operation were the NCEP/NCAR Reanalysis (Reanalysis-1) \cite{Kalnay96}, a global reanalysis of atmospheric data spanning 1948 to present and the NCEP/DOE Reanalysis (Reanalysis-2) project \cite{Kanamitsu02}, spanning 1979 to present. The Climate Forecast System Reanalysis (CFSR) \cite{Saha10} was an effort to generate an uniform, continuous, and best-estimate record of the state of the ocean-atmosphere interaction for use in climate monitoring and diagnostics. It is a global reanalysis, spanning data from January 1979 through March 2011. ERA-Interim \cite{ERA_Interim} is a global atmospheric reanalysis developed by the ECMWF. It covers from 1979, continuously updated in real time. The spatial resolution of the data set is approximately 15 km, on 60 vertical levels from the surface up to 0.1 hPa. ERA-Interim provides 6-hourly atmospheric fields on model levels, pressure levels, potential temperature and potential vorticity, and 3-hourly surface fields. ERA-5 \cite{ERA5} is the latest reanalysis project from the ECMWF. This reanalysis covers the Earth on a 30km grid and resolves the atmosphere using 137 levels from the surface up to a height of 80km. The Modern-Era Retrospective analysis for Research and Applications (MERRA) dataset was released in 2009 \cite{Rienecker11}. MERRA data span the period from 1979 through February 2016 and were produced on a $0.5^{\circ} \times 0.66^{\circ}$ grid with 72 layers. MERRA was used to drive stand-alone reanalyses of the land surface (MERRA-Land) and atmospheric aerosols (MERRAero). It is also worth mentioning the Japanese 25-year Reanalysis (JRA-25) \cite{JRA25} which is the first long-term global atmospheric reanalysis undertaken in Asia, and it covers the period 1979-2004.

There are also Regional reanalysis, such as the North American Regional Reanalysis (NARR)\cite{Mesinger06}. It is a regional reanalysis of North America containing temperatures, winds, moisture, soil data, and many other parameters. The high-resolution reanalysis system COSMO-REA6 has been developed based on the NWP model COSMO \cite{Bollmeyer15}. This is a regional reanalysis system for Continental Europe. This reanalysis data set currently covers the period 1995-2016 and it is currently in operation.
The European Reanalysis and Observations for Monitoring project (EURO4M) \cite{Euro4M} is a EU funded project that provides timely and reliable information about the state and evolution of the European climate.

Finally, note that there are several works focused on direct comparison of several reanalysis for the evaluation of different meteorological phenomena \cite{Hodges11,Bao12,Chaudhuri14}.


\section{Case studies on ML information fusion in Earth observation}

In this section we show empirical evidence of the performance of different ML fusion algorithms, working at different fusion levels in real practical EO problems. We discuss here four case studies: first, we present a problem of gap filling of several soil moisture time series from multiple microwave satellite data, working at different resolutions. In the second case study, we describe different algorithms that blend heterogeneous satellite sources in the optical range at different spatial and temporal scales in the Google Earth Engine platform. The third case study shows how natural hazards can be better predicted and modelled with ML data fusion. Finally, we introduce a case study where we discuss the fusion methodologies of EO and social media, stressing the importance and of unstructured data in EO, and the difficulty of their management.

\input{case_study_gapfilling.tex}
\input{case_study_assimilation.tex}
\input{case_study_natural_hazards.tex}
\input{case_study_socialmedia.tex}

\section{Conclusions, discussion and outlook}

In this article we have reviewed the state-of-the-art on Machine Learning (ML) information fusion for Earth observation (EO). The article, with a clear practical application, has been structured on literature review, data-sources and models description and some selected case studies where ML fusion information has obtained excellent performance in real problems. We have shown that the amount and diversity of the available data sources for EO has made information fusion a key step for successful real applications, with high societal, economical and environmental implications. We anticipate a huge impact in the upcoming years given the ever-growing increase, improvement, and ubiquity in sensory systems used in the EO field. The temporal, spatial and spectral resolutions are increasing dramatically, higher resolution models are now available, and social networks data promise to complement the view of the processes occurring in all spheres of the Earth system. We have also shown that ML approaches have the ability to blend and extract knowledge from these data, obtaining excellent results in a large number of EO problems and applications. 

There is an important challenge that the EO community has to face: the scalability of algorithms in the big data era. Many interesting approaches exist nowadays (high performance computing platforms, more efficient and sparse algorithms, green AI methodologies, etc.), and we expect much more advances on this in the near future. Besides this more technical problem, we can identify two important stepping stones: how to extract information in highly unstructured data, and how to achieve understanding through information fusion. On the one hand, the platforms and sources of information may vary considerably in multiple dimensions. For example, the types of properties sensed and the spatial and spectral resolutions of the data, and sometimes the temporal resolution of the corresponding sensor. This also applies even to the sensors that are mounted on the same platform, or that are part of the same satellite configuration. The rapid increase in the number and availability of data with an enormous amount of heterogeneity, and non-stationary properties, creates serious challenges for their effective and efficient processing. For a particular problem at hand, there may exist multiple remote sensing and ancillary datasets which leads to a dilemma: how best to integrate multiple datasets, probably unstructured and non-stationary, for maximum utility? It is currently one of the main challenges in EO. On the other hand, and perhaps more important, we have the issue of problem understanding through information fusion. Blending the information provided by different sources and ancillary systems not only can be an efficient way of improving the performance of ML algorithms in specific problems, but also a way of improving the understanding of the systems, usually as part of larger structures, and providing this way a better physical interpretation of the results obtained. The challenge is here to combine adequately sources of information which can provide very different perspectives to the problem, or even different possible interpretations depending on the case, and which improve the performance of the ML algorithms at the same time.

To tackle the previous challenges, we identify several exciting venues that open in the immediate future of ML information fusion in Earth sciences and EO in particular. A promising approach is about designing ML models that incorporate {\em domain knowledge} internally in a more sophisticated approach to data assimilation. Two methodological approaches have potential in facilitating the fusion of data-driven and arbitrary data sources and also physical models: probabilistic programming and differentiable programming. Probabilistic programming allows for accounting of various uncertainty aspects in a formal but flexible way, which allows in principle to account for data and model uncertainties, inclusion of priors and constraints coming from ancillary data and/or process-based (theory-driven) modeling. Differentiable programming, on the other hand, allows for efficient optimization of arbitrary losses owing to automated differentiation, which might help make the large, nonlinear and complex modeling more tractable. Attaining hybrid physics-aware ML will allow us advancing in improved modeling in terms of consistency and interpretability.

Understanding, especially from heterogeneous data, is harder than predicting, and ML algorithms are currently only mere (yet powerful) interpolation techniques: they excel in fitting arbitrary functional data relations but do not have a clear notion of the underlying causal relations. Despite the great predictive capabilities of current methods, there is still little actual {\em learning} in the ML information fusion pipeline. In this context, {\em causal inference} methods aim at discovering and explaining the causal structure of the underlying system. When interventions in the system are not possible, {\em observational causal inference} comes into play. Observation-based causal discovery aims at extracting potential causal relationships from multivariate datasets, and goes beyond the commonly adopted correlation approach, which merely obtains associations between variables. Today the science of ``causal inference'' is sufficiently advanced to unravel relations between multiple coupled variables beyond correlations even in the presence of non-linearities and non-stationarities. Observational causal inference could make a decisive change in the way we process, analyze and understand multi-source and multi-sensory data.

\section*{Acknowledgments}
This research has been partially supported by the Ministerio de Econom\'{i}a y Competitividad of Spain (Grant Ref. TIN2017-85887-C2-2-P). GCV work is supported by the European Research Council (ERC) under the ERC-CoG-2014 SEDAL Consolidator grant (grant agreement 647423).

\end{document}

%% file: case_study_gapfilling.tex
\subsection{Multitemporal and multisensor gap-filling in remote sensing}

The presence of gaps in EO data limits their applicability in a number of applications that need continuous data. Standard techniques for gap filling temporal series such as linear or cubic interpolation, or auto-regressive functions fail to reconstruct sharp transitions or long data gaps. Also, they are not able to infer information from other collocated sensors measuring the same biophysical variable, which is the setting found, for instance, when harmonizing data from multiple satellites into consistent climate data records of Essential Climate Variables (ECVs)~\cite{GCOS}. Another challenging setting for standard gap-filling approaches is the fusion of collocated microwave and optical observations for cloud-free estimates of vegetation descriptors, which needs to exploit the relationships between the two~\cite{Pipia19}. In this section, we show how we can efficiently deal with the spatio-temporal gaps of collocated satellite-based observations by employing a multi-output Gaussian Process model based on the Linear Model of Corregionalization (LMC)~\cite{Alvarez11}. The method allows learning the relationships among the different sensors and build an across-domain kernel function to transfer information across the time series and do predictions with associated confidence intervals on regions where no data are available. 

\begin{figure}[t!] \centering
    \includegraphics[width=0.95\textwidth]{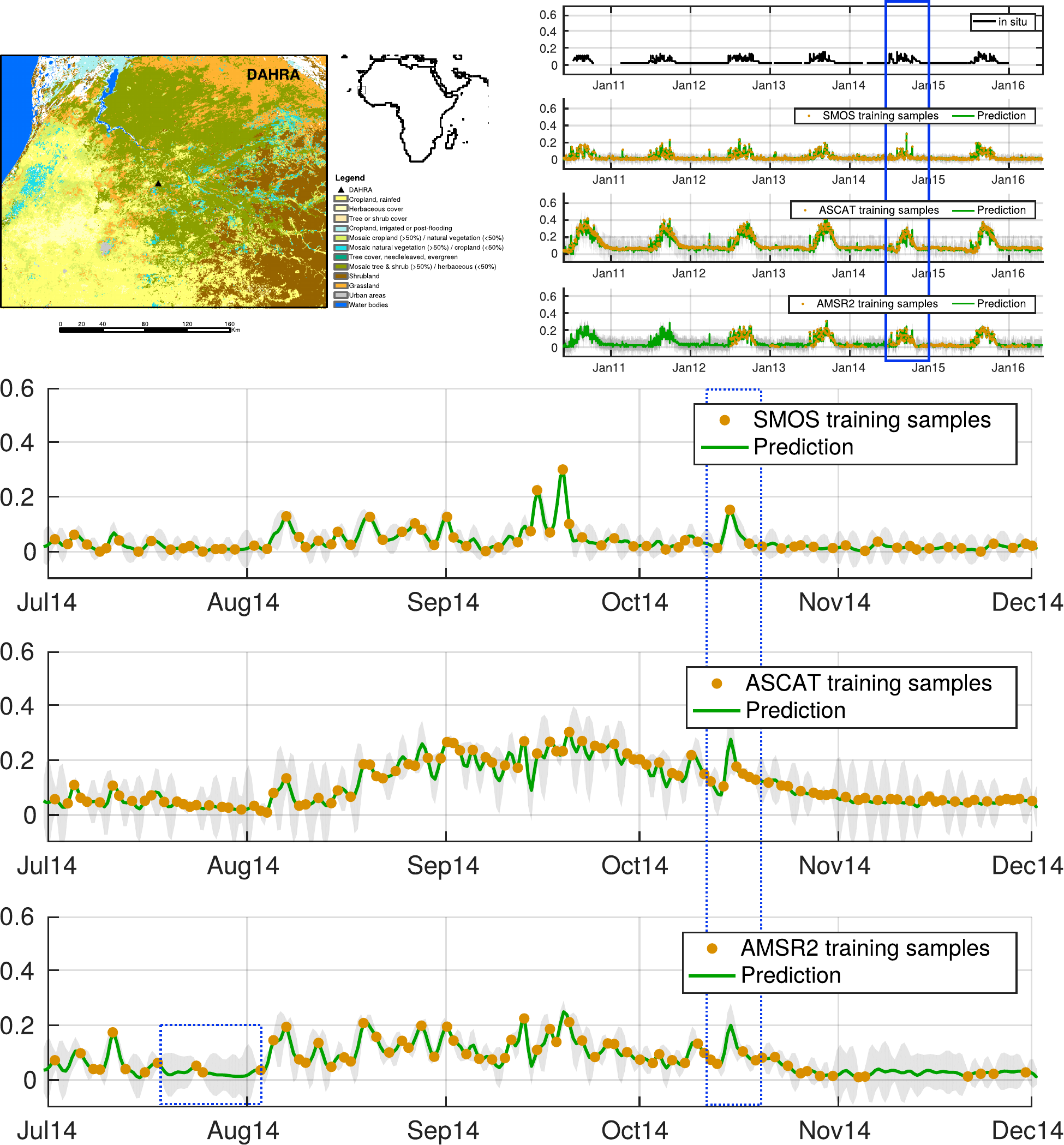}
    \caption{Results of the application of the multi-output LMC-GP gap-filling technique at DAHRA validation site (1 station). Top left: site location and land use map. Top right: time series of \textit{in-situ} (black-lines), and satellite-based soil moisture estimates from SMOS, ASCAT and AMSR2 (orange dots denote the training data and green lines the predictions). The blue rectangle indicates the time period that is represented in the bottom figure. The bottom-left dashed rectangle exemplifies how the method reconstructs long data gaps in AMSR2 based on no-rain information from the other two sensors, assigning a higher uncertainty when no training data is available. The bottom-right dashed rectangle points out a specific rainfall event that was captured only by SMOS and is accounted for in the reconstruction of ASCAT and AMSR2 time series.}
    \label{fig:dahra}
\end{figure}

We illustrate the procedure using soil moisture time series from three spaceborne microwave sensors, which are integrated in the ESA Climate Change Initiative (CCI) soil moisture product \cite{Dorigo2017}: the ESA's SMOS L-band radiometer, the AMSR2 C-band radiometer on-board JAXA's GCOM-W1, and the ASCAT C-band scatterometer on-board Eumetsat MetOp satellites. The temporal period of study is 6 years, starting in June 2010. Each product presents different observational gaps due to the presence of Radio Frequency Interferences at their operating frequency or high uncertainty in their inversion algorithm (e.g. presence of snow masking observations, dense vegetation, or high topography). The problem we face here is that we need a gap-filling methodology able to handle several outputs together and force a ``sharp'' reconstruction of the time series so that fast dry-down and wetting-up dynamics are preserved (avoid smoothing). We show the application of the LMC multi-output GP regression at two {\em in-situ} soil moisture networks from the \href{https://ismn.geo.tuwien.ac.at}{International Soil Moisture Network}: REMEDHUS in Spain (17 stations~\cite{Sanchez12}), and DAHRA in Senegal (1 station~\cite{dahra}). In terms of temporal coverage, they are representative of best-case (REMEDHUS), and wort-case (DAHRA) scenarios, where best satellite coverage is of 91\% and of 64\% of the study period, respectively. 

Results of the application of the proposed LMC-GP over DAHRA are shown in Figure ~\ref{fig:dahra}, together with the original satellite time series and the {\em in-situ} data as a benchmark. It can be seen that the reconstructed soil moisture time series follow closely the original time series, capturing the wetting-up and drying-down events and filling the missing information (e.g. see the peak in October 2014 which was captured only by SMOS and is reproduced by the three reconstructed time series). Also, predictions have associated uncertainties related to the availability of training data for each specific sensor. It is remarkable that for AMSR2 the reconstructed time series back-propagate to dates before the satellite was launched (shown here for illustration purposes), yet they look very consistent with the real satellite data. Given the soil moisture products present no-data in different time and space locations, the method allows to provide predictions at all time stamps where at least there is one satellite measurement, maximizing therefore the spatio-temporal coverage of the data sets.

\begin{table}[!ht]
\small
\caption{Mean error (ME) [m$^3$m$^{-3}$], unbiased RMSE (ubRMSE) [m$^3$m$^{-3}$] and Pearson's correlation (R) for the original and reconstructed satellite time series against {\em in-situ} measurements from REMEDHUS and DAHRA networks. Variable `gaps' reports the percentage of days that were gap-filled in the reconstructed series.}
\vspace{-0.5cm}
\begin{center}
\small \renewcommand{\tabcolsep}{0.1cm}
\begin{tabular}{c|cccc|cccc}
\toprule
 & \multicolumn{4}{c|}{\bf{REMEDHUS}} & \multicolumn{4}{c}{\bf DAHRA } \\
& ME & ubRMSE & R & gaps[\%] & ME & ubRMSE & R &  gaps[\%] \\
\midrule
 SMOS & -0.032 & 0.003 & 0.81 & - & -0.0143 & 0.001 & 0.79   & - \\
SMOS rec & -0.033 & 0.003 & 0.81 & 8.58  & -0.014 & 0.001 & 0.78   & 54.14 \\
\midrule
 ASCAT & 0.002 & 0.004 & 0.79 & -  & 0.071 & 0.002 & 0.70   & - \\
 ASCAT rec & -0.001 & 0.004 & 0.78 & 23.68 & 0.064 & 0.002 & 0.70   & 36.15 \\
\midrule
 AMSR2 & 0.118 & 0.005 & 0.86 &  - & 0.026 & 0.002 & 0.73   & - \\
 AMSR2 rec & 0.084 & 0.005 & 0.81 &  52.24 & 0.019 & 0.001 & 0.79 &  77.26 \\
\bottomrule
\end{tabular}
\end{center}
\label{table:sm_stats}
\end{table}

Statistical scores from comparison with {\em in-situ} data at the two sites of the original and reconstructed time series are shown in Table~\ref{table:sm_stats}. The analyses show that Pearson's correlation coefficient (R), mean error (ME) and unbiased root-mean-squared error (ubRMSE) with respect to {\em in-situ} data remain within reasonable bounds and are not affected to a high degree by the reconstruction, even in the more extreme case of DAHRA. These results provide confidence in the proposed technique and their potential to mitigate the effect of missing information in satellite-based observational records.

%% file: case_study_assimilation.tex
\subsection{Modern data assimilation and hybrid modeling in geosciences}

In this case study, we focus on image data fusion (blending) methods as optimal approaches for overcoming individual sensor's limitations and combining different multiresolution datasets. Blending Landsat and MODIS has been the preferred sensor combination in the literature and enabled predicting gap free surface reflectances \cite{gao2006blending,gevaert2015comparison} at Landsat spatial resolution (30 m). Both missions provide long time series of data with a high degree of consistency. The MODIS sensor, on board of Terra and Aqua platforms, provides global observations and a daily revisit cycle at a cost of  having a coarse spatial resolution (250-1000m). This resolution clearly limits its utility for fine-scale environmental applications, but on the other hand, MODIS high temporal resolution allows tracking rapid land-cover changes while maximizes the possibility of having cloud-free observations. We have also capitalized on using these two sensors, especially because the proposed approach has been specifically designed to exploit past temporal information to improve the results. 

Here, we focus on a KF logic method named HIghly Scalable Temporal Adaptive Fusion Model (HISTARFM). This algorithm was implemented in the Google Earth Engine (GEE) cloud computing platform \cite{gorelick2017google} and consists of a bias-aware Bayesian data assimilation scheme \cite{dee1998data}. HISTARFM uses two Kalman estimators operating simultaneously to reduce the amount of noise and decrease possible biases (if present) in the predicted Landsat spectral reflectances. The first estimator is an optimal interpolator (a special case of the Kalman filter with no dynamic model) that produces estimates of Landsat reflectance values for a given time by combining a Landsat climatology (mean monthly values considering many years) and linearly blended Landsat and MODIS spectral information. The second coupled estimator is an additional Kalman filter which is in charge of correcting dynamically possible biases of the reflectance produced (if present) by the first estimator. The next figure shows an example of the results of the HISTARFM algorithm over an area with massive gaps due to cloud contamination and sensor malfunctioning.

\begin{figure*}[h!]
\subfigure[Original RGB composite with Landsat data ]{\includegraphics[width=.487\textwidth]{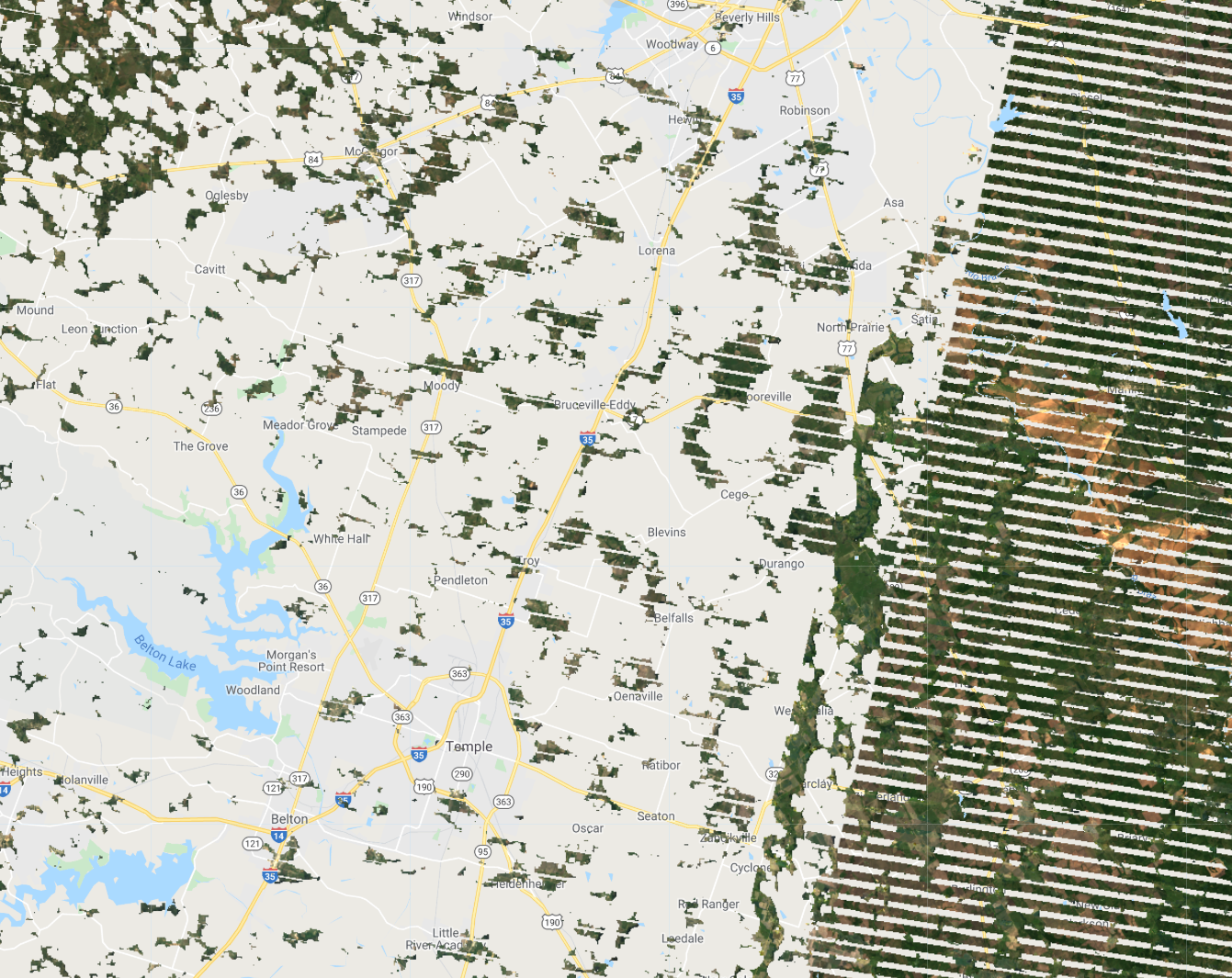}\label{fig:twitter}}
\subfigure[HISTARFM gap filled RGB composite]{\includegraphics[width=.49\textwidth]{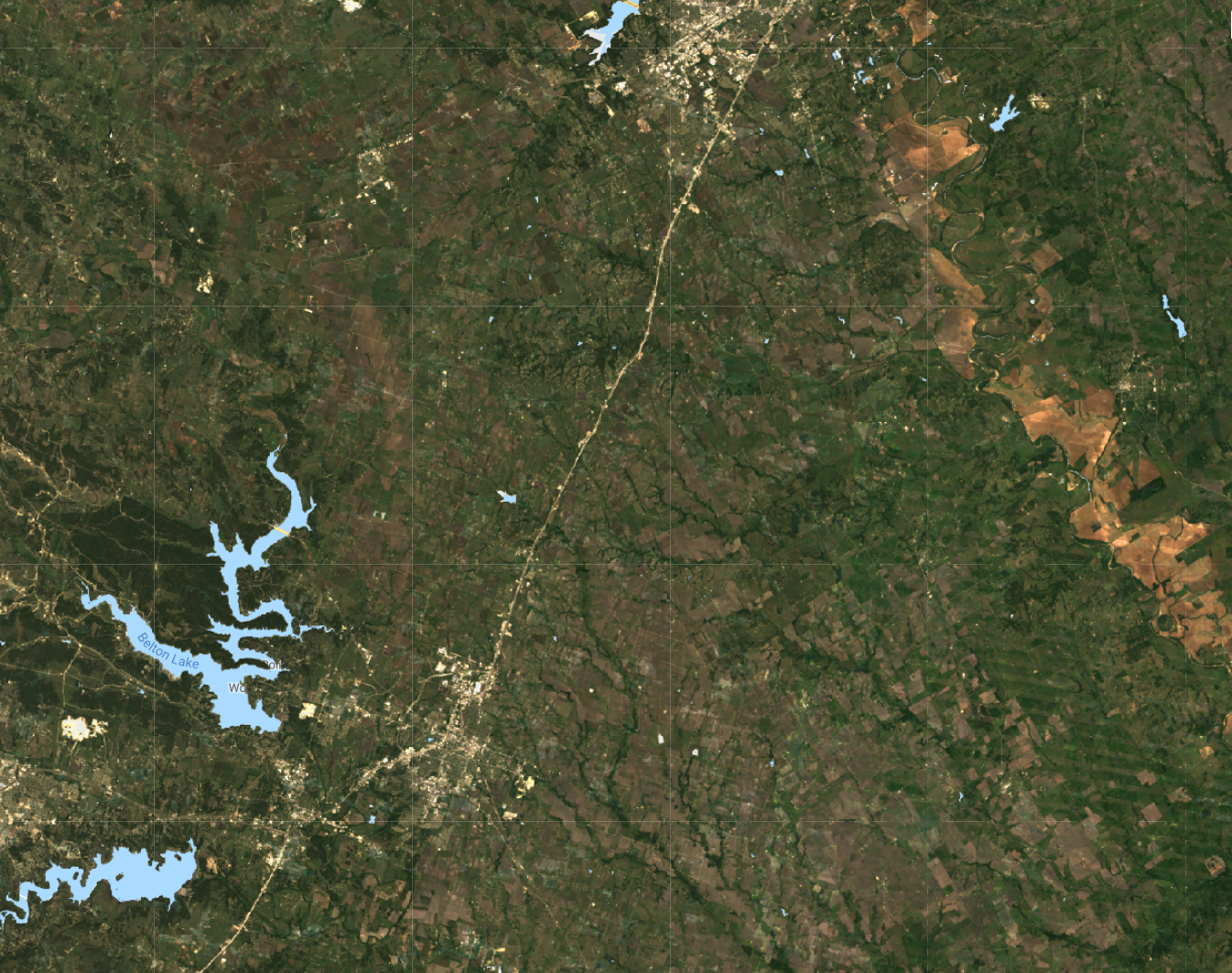}\label{fig:twitter-sentiment}}
\caption{Differences between the original Landsat data and the gap filled data set processed with the the proposed data assimilation approach. Both images correspond with a  cropland area in Texas state (US) for the date May 2010.}
\end{figure*}

HISTARFM takes advantage of the GEE  platform, this enables to process huge amounts of data significantly faster than other approaches available in the literature. The validation of the proposed method over 1050 sites spread out over the conterminous United States indicated the feasibility of the method and provided satisfactory results. The relative mean errors remained below 2$\%$ (in all spectral bands) and the relative mean absolute errors and relative root mean squared errors ranged between low to moderate (10-20$\%$) depending on the spectral band. Moreover,  the high degree of agreement between the validation errors and the predicted uncertainties by HISTARFM indicated the utility of this information for error propagation purposes. 

%% file: case_study_natural_hazards.tex
\subsection{Natural hazard prediction and data fusion}

Natural hazards cause thousands of deaths and inflict tremendous societal damage every year. To demonstrate the catastrophic influence of such hazards, between 2005 and 2014, 700,000 people were killed, and 1.7 billion people were affected worldwide by disasters\footnote{https://www.unisdr.org/}. This clearly demonstrates the importance of developing accurate and efficient mapping, modeling, and prediction techniques to reduce the catastrophic impact of natural hazards. 

The success of information fusion techniques has revolutionized the challenging and vitally important applications of modeling natural hazards \cite{AP1}. In this context, the fusion of multiple sources of data using ML techniques has been reported as an effective tool to greatly contribute to increasing the accuracy of the prediction models \cite{AP2, AP3,P20,P21}. 

To provide a bird's-eye view over the variety of data fusion models for hazard modeling, a number of relevant research studies have been summarized in Table \ref{table:Hazard}, which provides an overview of the key contributions, investigated data sources, and the tackled hazard type. The fusion of the multiple data sources, e.g., satellite imaging, radar data, laser point clouds, UAV images, weather stations, crowdsourcing data, social media, and GIS have shown their advantages to greatly enhance the robustness and performance of hazard detection and avoidance systems, leading to a safer planetary anytime, anywhere.
\begin{table}[!ht]
\scriptsize 
\vspace{12pt}
\caption{Notable Research studies on data fusion techniques for hazard modeling .}

\centering
\begin{tabular}{| p{3.0cm} |   p{5.5cm} | p{5.0cm}| p{1.4cm}|}
\hline
 \textbf{Research Studies}   &  \textbf{Contribution}  &  \textbf{Data fusion}    & \textbf{Hazard type} \\ 
\hline

Shi et al. (2019) \cite{AP4} & An enhanced flexible spatiotemporal data fusion model for prediction &	Fusion of Landsat and MODIS satellite data &	 Landslide  \\ 
\hline

Shanet al. (2019) \cite{AP5} & High-rate real-time GNSS seismology and early warning system &	Fusion of displacement and acceleration seismology data	& Earthquake \\
\hline

Yang et al. (2019) \cite{AP6} & Susceptibility mapping using the B-GeoSVC model and Hierarchical Bayesian method &	Fusion of regional and local information &	 Landslide  \\
\hline

Feng et al. (2019) \cite{AP7} & Integration of remotely sensed drought factors for accurate drought prediction &	Fusion of thirty remotely-sensed drought factors from the Tropical Rainfall Measuring Mission (TRMM) and MODIS &	 Drought \\
\hline

Zou et al. (2019) \cite{AP8} & Wildfire smoke simulations and observations for regional hazard prediction &	Fusion of PM2.5 air pollution &	 Wildfire \\
\hline

Knipper et al. (2019) \cite{AP9} &
Evapotranspiration prediction for irrigation management &	 Fusion of evapotranspiration time-series retrievals from multiple satellite platforms &	 Drought \\
\hline

Guerriero et al. (2019) \cite{AP10} &
Flood hazard mapping in convex floodplain: Multiple probability models fusion, bank threshold, and levees effect spatialization	& Fusion of LiDAR-derived high-resolution topography and ground-based measurements &	 Flood \\
\hline

Lee and Tien (2018) \cite{AP11} & Probabilistic Framework for disaster prediction  &	 Fusion of multiple sources, including physical sensors measuring environmental quantities and big data from social sensors &	 Disaster \\
\hline

Azmi and Rüdiger (2018) \cite{AP12} & Validating the data fusion‐based drought index (DFDI) and recalibrating the regional drought thresholds for increasing the predictive accuracy. &	Fusion of drought index &	 Drought \\
\hline

Alizadeh and Nikoo (2018) \cite{AP13} & A fusion-based methodology for drought prediction &	Fusion of diverse remotely sensed data &	 Drought \\
\hline

Pastick et al. (2018) \cite{AP14} & Spatiotemporal analysis of dryland ecosystems &	Data fusion of Landsat-8 and Sentinel-2, and MODIS &	 Drought \\
\hline

Li and Fan (2017) \cite{AP15} & Accurate landslides hazard prediction &	Fusion of remote sensing data and UAV &	 Landslide \\
\hline

Zhuo and, Han (2017) \cite{AP16} & Accurate drought prediction &	 Fusion of remotely sensed data and weather stations &	 Drought \\
\hline

Rosser et al. (2017) \cite{AP17} & Rapid and accurate flood inundation mapping 	& Fusion of social media, remote sensing, and topographic information & 	 Flood \\
\hline

Renschler and Wang (2017) \cite{AP18} &  Accurate flood prediction &	Fusion of multi-source GIS, hydraulic modeling based on remote sensing data, and LiDAR	& Flood \\
\hline

Hillen (2017) \cite{AP19} & Hazard prediction and mitigation &	Fusion of geo-information and mobility data 	& Geohazards and flood \\
\hline
\end{tabular}
\label{table:Hazard}
\vspace{24pt}
\end{table} 
 
Among the variety of geohazards, flood and drought modeling and prediction is regarded as a very complex phenomenon which is known to be among the least understood natural hazards due to its multiple causing reasons or contributing factors operating at different temporal and spatial scales. Fortunately, the application of data fusion in flood and drought modeling has evolved significantly compared to other hazards \cite{AP19,AP20,AP21}. To demonstrate the effectiveness of ML-based information fusion techniques for the challenging application of flood prediction, in the following, we provided a dedicated case study.

\subsubsection{Flood prediction by integrating remote sensing and weather station data}

Floods are increasingly recognized as a frequent natural hazard worldwide. Increasing the accuracy of flood susceptibility mapping is of utmost importance for efficient land use management, policy analysis, and the advancement of the mitigation structures to optimally reduce the devastation. 
\textbf{Introduction to flood susceptibility mapping: }
Hazard susceptibility mapping of flood is essential for mitigation due to their higher destructive power in a short period. ML-based methods are among the most popular methods used for accurate mapping of the flood hazard \cite{AP25}. Several comparative studies in the literature reported promising results using ML methods \cite{AP26, AP27}. Along with employing the hybridization and ensemble techniques for improving the accuracy of ML models, the data fusion techniques have shown promising results. The aim of this case study is to demonstrate the performance of a data fusion approach to integrate information obtained by weather stations, land survey, and satellite data to improve the accuracy of the flood superstability mapping.

\begin{figure*}
\centering
  \includegraphics[width=0.5\linewidth]{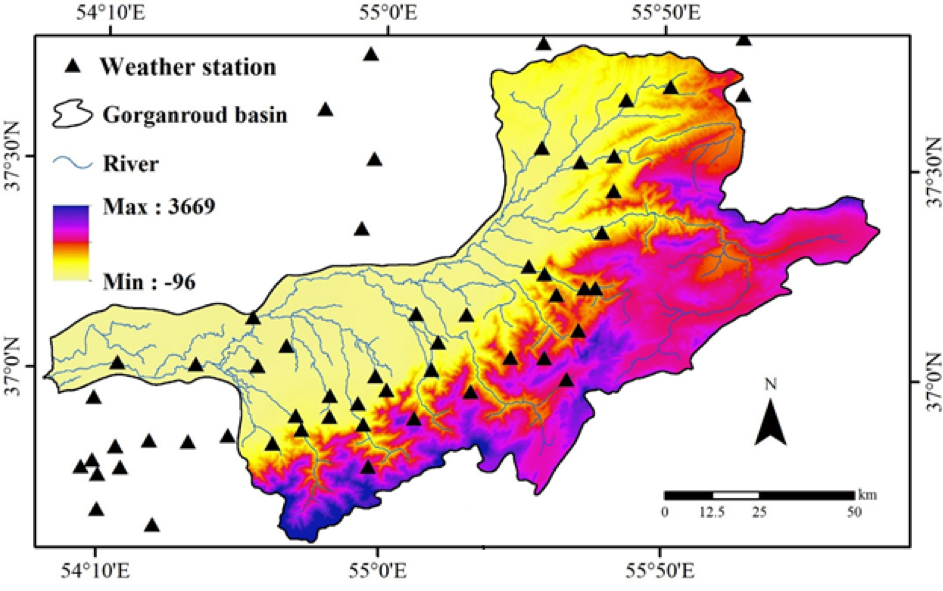}
\vspace{-1.2em}
  \caption{Location of the case study; distribution of the weather stations in the basin.}
\label{fig:statAP2}
\end{figure*}

\begin{figure*}
\centering
  \includegraphics[width=0.7\linewidth]{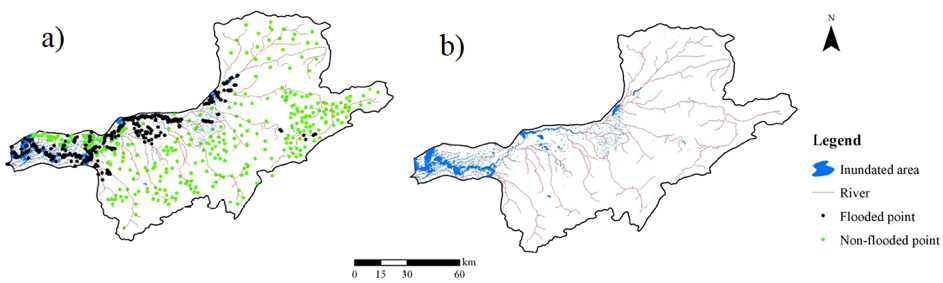}
\vspace{-1.2em}
  \caption{a) Flooded and non-flooded points (b) inundated regions according to Sentinel-2.}
\label{fig:statAP3}
\end{figure*}

\textbf{Materials and methods:}
The study area is Gorganroud Basin, located in the northwest of Iran between latitudes of 36º 25' to 38º 15' N and longitudes of 56º 26' to 54º 10' E. In this case study, the data fusion is conducted by locating the flooded and non-flooded points and identifying the inundated regions using Sentinel-2 satellite images. Due to the lack of recorded location of flood occurrences, the inundation areas are identified using the Modified Normalized Difference Water Index (\red{MNDWI}) of Sentinel-2. Radiometrically calibrated and terrain corrected Sentinel-2 Level-1C dataset is stored within GEE to support the free cloud computing facilities for this case study \cite{AP28}. Fig. \ref{fig:statAP2} shows the study area. The inundated area is extracted by MNDWI during a period from March and April of 2019 when the flood affected the region (Fig. \ref{fig:statAP3}). Furthermore, feature selection using simulated annealing and modeling through RF is used to identify the hazard areas. The validation is done using hit and miss analysis.

After identifying the inundated area, the number of 368 flash-flood locations were randomly considered from the inundated points and their locations were confirmed through field surveys. For modeling, the inundated points were considered as the dependent variable and used for modeling with RF. 

\begin{figure*}
\centering
  \includegraphics[width=0.7\linewidth]{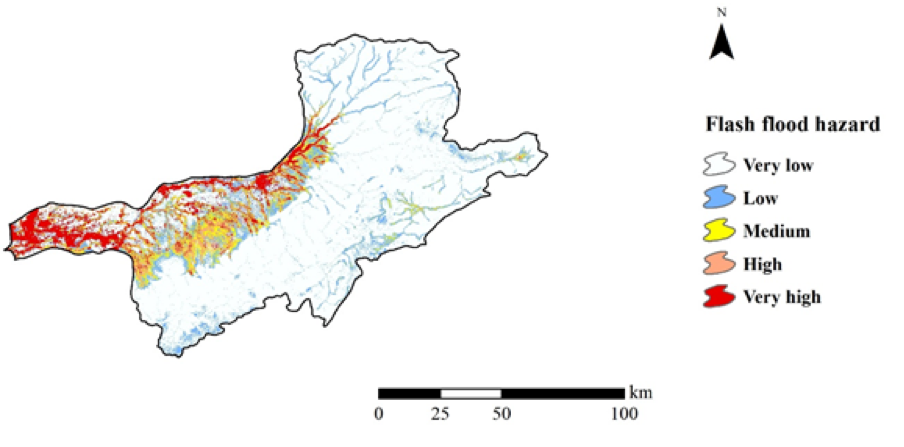}
\vspace{-1.2em}
  \caption{Hazard sustainability mapping of flood using RF.}
\label{fig:statAP4}
\end{figure*}

\textbf{Modeling:} 
After preparing the predictand flood/non-flood locations as input and the predicting variables as output, the model is developed where the values of 0 and 1 were assigned to the non-flood and flood occurrence locations, respectively. From the whole dataset, 70\% of the data is considered for training while the remaining 30\% of the data is used for testing. A 10-fold cross-validation methodology was used to train the ML models. The results of the hazard modeling using RF is shown in Fig. \ref{fig:statAP4}. 

%% file: case_study_socialmedia.tex
\subsection{Fusion of Earth data and Social media}

Earth data is routinely used to infer many aspects of Earth. Since most Earth data are generated using satellites or airborne platforms, they are often limited to a birds-eye perspective. Given the high sensor quality and non-invasiveness of remote sensing, such birds-eye Earth data have become a primary driver for understanding the morphological structure of Earth, including applications ranging from biology \cite{buchanan2009delivering,giri2011status}, environmental and social sciences \cite{skidmore2003environmental,turner2003methodological}, cartography and mapping \cite{huang2018urban}, among others. Unfortunately, this non-invasiveness of the birds-eye perspective implies some limitations in the semantic concepts one can distinguish. This difference is manifested in the distinction of land-cover mapping from land-use mapping or as well between building function (e.g., what people are supposed to do in this location) and activity estimation (e.g., what people are observed to do in reality). While the first one is what a satellite or airborne platform can sense, the second one is even more important to many applications. In this context, additional data and measurements can greatly help resolve these ambiguities. Such additional (covariate) data sources include base map data including information like building footprints, street networks, cadastral information, historical information or and data contributed by citizens, for example, through using social media. Social media information is envisioned to augment the birds-eye view with ground-level features like images or text originating from a certain location or more abstract anthropogenic signals correlated to population density, wealth, or other spatial distributions of interest.

In general, there are three categories of how location-based social media can be analyzed and fused with ESS data: one is based on the metadata. In computer science, metadata comprises all data that describes the data at hand, usually including information on time, identity, place, size, and user profiles. The second category of exploiting social media is based on text mining, trying to understand and relate the message content to the Earth. Finally, the third category considers social media images and video sequences. Similarly to the text case, many images in social networks do not relate to the location they are sent from. Hopefully, the totality of images and videos in a certain area tells something about the area or one can detect which messages are related to a given location. All three aspects of social media require different data mining, ML and data fusion approaches as they all pose very different challenges to the overall system.

\subsubsection{Spatial Statistics of Metadata}

\begin{figure*}[!ht]
\subfigure[Global Twitter Density]{\includegraphics[width=.49\textwidth]{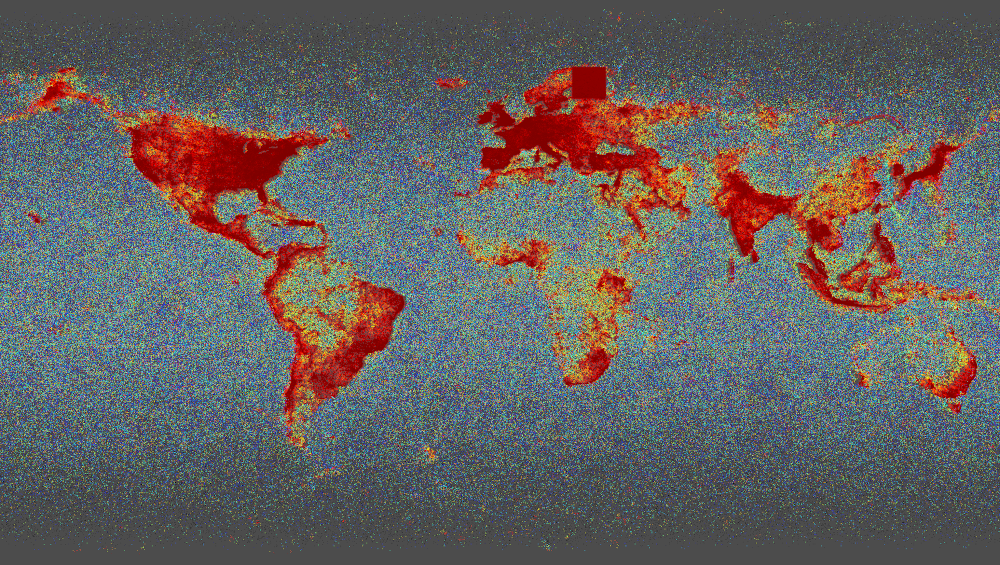}\label{fig:twitter}}
\subfigure[Positive Sentiment of Twitter messages around New York]{\includegraphics[width=.49\textwidth]{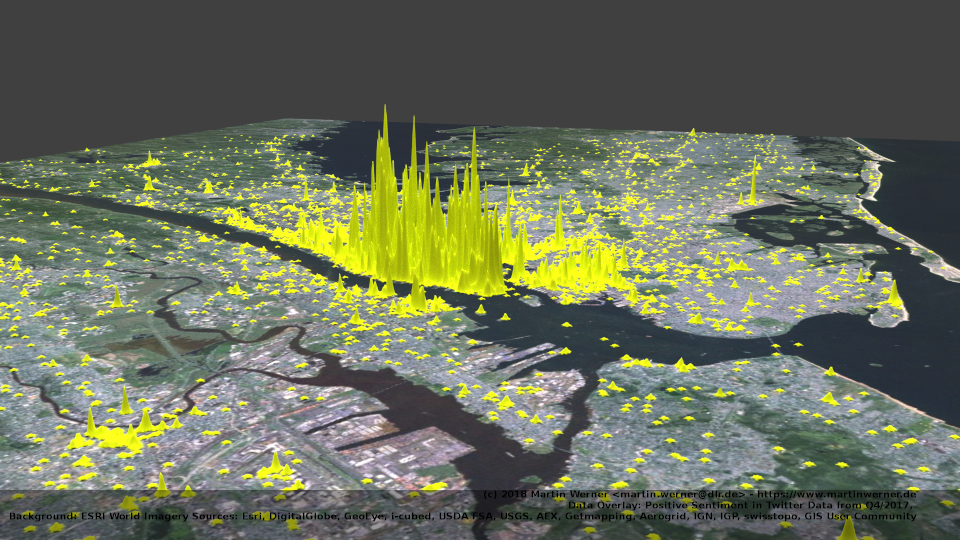}\label{fig:twitter-sentiment}}
\caption{Illustrations of Social Media Statistics on the example of Twitter public stream data.}
\end{figure*}

The most traditional technique relies on spatial summary statistics of messages. One assumes that the patterns in which messages are generated are related to interesting factors which influence ESS parameters of interest. For example, social media users are typically quiet while sleeping and have a detectably higher activity during the day \cite{zheng2018survey}. That is, messages that originate from a certain location late in the evening as well as in the early morning with a break of a few hours might indicate some home or sleep location, definitely a spatial information. Fusing with overhead imagery, we might be able to distinguish between a hotel (large building, social media activity very low during late night) and a residential building (small building, much green around, social media activity low during late night). In addition, the type of user or intention of the message is reflected in metadata to some extent. Marketing and job announcements of corporations, for example, tend to use the full capacity of social media containing as many hashtags as possible to reach many users, and contain many words exploiting the full capacity of the message. It has been shown that simple features extracted from social media metadata alone can increase the convergence speed and final quality of a deep learning model predicting urban land use modeled after local climate zones paradigm by a significant margin \cite{Leichter18}. Figure \ref{fig:twitter} depicts spatial density of Twitter data measured as the radius of a $k$-nearest-neighbor environment of each point. As can seen, Twitter usage is tightly connected to many socio-demographic factors. This relation has been discussed in \cite{li2013spatial} as well.

\subsubsection{Text Mining}

Text mining comprises a set of techniques in which natural language text is transformed into a numeric representation that captures some semantics of the text. Many techniques have been introduced in the past that are based on character and word frequencies including TF-IDF ranking of documents in keyword search \cite{lott2012survey}, or sparse text mining based on word-document matrices including topic modelling and Latent Dirichlet Allocation (LDA) \cite{chen2019experimental}. However, the limitation of these methods has usually been to find a good vocabulary for a given task that does not contain meaningless words and pre-processing text in order that different forms of words are detected as the same word in these methods. When it comes to spatial data, however, language is gradually changing with location and might jump to a completely different tongue when crossing borders, complicating this process even more. In the deep learning area, text embeddings have become feasible which are a bit less dependent on these basic, but very hard language pre-processing techniques, and exploit very large datasets to cover the variations automatically \cite{manna2019effectiveness,tang2014learning}. These map a vocabulary into a Euclidean space of chosen dimension in a way such that semantically similar words end up near each other. Doing this in a certain way leads to the surprising property that the addition in the vector space becomes semantically sensible with some limitations \cite{Boja16}. But generally, one aims to become able to perform {\em semantic operations} in the latent space as follows:
\begin{align*}
\text{King - Man + Woman} & \approx \text{Queen}\\
\text{Paris - France + Germany} & \approx \text{Berlin}
\end{align*}
Such numeric representations have then been used in deep learning using long short-term memories (LSTMs) \cite{wang2015unified} in order to allow for the deep learning system to analyze sentences taking care to important semantic words like negations.  In the last months, advanced language models based on transformer architectures like BERT and GPT2 have shown impressive language understanding and generation capacities. Their careful application to social media in the context of ESS is a promising direction for future research. In relation to ESS, text embeddings have been successfully applied to distinguish residential and commercial buildings \cite{igarssbtclassification}. In addition, standard challenge problems of text mining such as sentiment analysis can as well bring interesting spatial information to life. Sentiment analysis has been well-researched in the natural language domain and in the computer science and ML domain in the context of learning from sparsely labeled streams \cite{iosifidis2019sentiment}. The idea is to assign a rating ranging from negative to positive to a text capturing the emotion it represents. Figure \ref{fig:twitter-sentiment} depicts the distribution of positive sentiment tweets around New York, which seems to be somehow skewed towards commercial centers.

\subsubsection{Image Mining and Multimedia Analysis}

In a similar vein to text mining, the whole body of image analysis and computer vision research can be applied to social media images too. This can help collect information about the Earth surface from a ground-level view, assuming that there are enough images taken at the location from where they are sent. Similar to the text case, images attached to social media messages might not be related to the location where they are posted. Consequently, the multimedia information available from social media is extremely noisy and one must take this into account \cite{igarssmutualinformation}.

Social media images have been used to extract statistics such as object counts essentially augmenting message metadata and opening the field for applications of spatial statistics. However, images can as well be directly used in order to classify, for example, land-use classes \cite{dlr126371,groundaerialfusion}. In this case, however, causal relations usually break and the system will be ``right for the wrong reason'', a typical side-effect of overfitting. In general, the social media stream will contain images that might not be representative for the actual context of the post, where this context information can be the physical surroundings as well as the activity (e.g., dining) or the socio-demographic context (e.g., rich vs. poor). An end-to-end ML system can learn to exploit all various types of context in the images and this abstract context is related to the location of origin at least in the sense that a user has been in a given context in this location. Consequently, it is not surprising that social media images -- though they look like they do not relate significantly to the surroundings as they often do not depict the surroundings -- reveal a lot of social, spatial, and economic information and can, therefore, help augment typical ESS questions \cite{cao2018integrating}.

\subsubsection{Summary}

Social media comprises a nice yet challenging data source for Earth observation. While it is obvious that social media correlates to human activities and is dense in urban regions, it is surprising that the content of the messages and their metadata can be used to distinguish traditional land cover and land use ambiguities in remote sensing imagery. However, methods of learning from such noisy data streams with such sparse labels are still in its infancy and need additional breakthroughs in unsupervised ML, natural language processing, and spatial data science in order to provide full potential.

%% file: Infus_ESS_revised.bbl
\begin{thebibliography}{1}

\bibitem{Kump04}
L. R. Kump, J. F. Kasting, R. G. Crane, The Earth system, Pearson Prentice Hall, 2004.

\bibitem{Stanley05}
S. M. Stanley, Earth system history, Macmillan press, 2005.

\bibitem{Brasseur05}
G. P. Brasseur, S. Solomon, Aeronomy of the Middle Atmosphere: Chemistry and Physics of the Stratosphere and Mesosphere, Springer, 3rd Edition revised and enlarged, 2005.

\bibitem{Jacobson00}
M. Jacobson, R. J. Charlson, H. Rodhe, G. H. Orians, Earth System Science: From Biogeochemical Cycles to Global Changes, Academic Press, 2000.


\bibitem{Wald00}
L. Wald, A conceptual approach to the fusion of Earth observation data, Surv. Geophys., 21 (2000) 177-186.

\bibitem{Cherkassky06}
V. Cherkassky, V. Krasnopolsky, D. P. Solomatine, J. Valdes, Computational intelligence in Earth sciences and environmental applications: Issues and challenges, Neural Net., 19(2) (2006) 113-121.

\bibitem{Torabzadeh14}
H. Torabzadeh, F. Morsdorf, M. E. Schaepman, Fusion of imaging spectroscopy and airborne laser scanning data for characterization of forest ecosystems -- A review, ISPRS J. Photogram. Remote Sens., 97 (2014) 25-35, 2014.

\bibitem{Chen15}
B. Chen, B. Huang, B. Xu, Comparison of spatiotemporal fusion models: A review, Remote Sens., 7(2) (2015) 1798-1835.

\bibitem{Gomez15}
L. G\'omez-Chova, D. Tuia, G. Moser, G. Camps-Valls, Multimodal Classification of Remote Sensing Images: A Review and Future Directions, Proc. IEEE, 103(9) (2015) 1560-1584.

\bibitem{Schmitt16}
M. Schmitt, X. X. Zhu, Data Fusion and Remote Sensing: An ever-growing relationship, IEEE Geosci. Remote Sens. Mag., 4(4) (2016) 6-23.

\bibitem{Ghassemian16}
H. Ghassemian, A review of remote sensing image fusion methods, Inf. Fus., 32 (2016) 75-89.

\bibitem{Garzelli16}
A. Garzelli, A review of image fusion algorithms based on the super-resolution paradigm, Remote Sens., 8(10) (2016) 797.

\bibitem{Yokoya17}
N. Yokoya, C. Grohnfeldt, J. Chanussot, Hyperspectral and multispectral data fusion: A comparative review of the recent literature, IEEE Geosci. Remote Sens. Mag., 5(2) (2017) 29-56.

\bibitem{PG22}
P. Ghamisi, J. Plaza, Y. Chen, J. Li, A. Plaza, Advanced Supervised Classifiers for Hyperspectral Images: A Review, IEEE Geosci. Remote Sens. Mag., 5(1) (2017) 1-7.

\bibitem{Gham19}
P. Ghamisi, B. Rasti, N. Yokoya, Q. Wang, B. Hofle, L. Bruzzone, et al., Multisource and multitemporal data fusion in remote sensing: a comprehensive review of the state of the art, IEEE Geosci. Remote Sens. Mag., 7(1) (2019).

\bibitem{Guo16}
H. Guo, L. Wang, D. Liang, Big Earth Data from space: a new engine for Earth science, Sci. Bull., 61(7) (2016) 505-513.

\bibitem{Gibert18}
K. Gibert, J. S. Horsburgh, I. N. Athanasiadis, G. Holmes, Environmental Data Science, Env. Model. Soft., 106 (2018) 4-12.

\bibitem{Ball17}
J. E. Ball, D. T. Anderson, C. S. Chan, Comprehensive survey of deep learning in remote sensing: Theories, tools and challenges for the community, J. App. Remote Sens., 11(4) (2017) 042609.

\bibitem{Ma19}
L. Ma, Y. Liu, X. Zhang, Y. Ye, G. Yin, B. A. Johnson, Deep learning in remote sensing applications: A meta-analysis and review, ISPRS J. Photogram. Remote Sens., 152 (2019) 166-177.


\bibitem{PLi2019}
S. Li, W. Song, L. Fang, Y. Chen, P. Ghamisi and J. A. Benediktsson, Deep Learning for Hyperspectral Image Classification: An Overview, in IEEE Trans. Geosci. Remote Sens., 57(9) (2019) 6690-6709.

\bibitem{Yuan20}
Q. Yuan, H. Shen, T. Li, Z. Li, S. Li, Y. Jiang, et al., Deep learning in environmental remote sensing: Achievements and challenges, Remote Sens. Env. 241 (2020) 111716.

\bibitem{Zhang18}
Q. Zhang, L. T. Yang, Z. Chen and P. Li, A survey on deep learning for big data, Inf. Fus., 42 (2018) 146-157.

\bibitem{Reichstein19}
M. Reichstein, G. Camps-Valls, B. Stevens, M. Jung, J. Denzler, N. Carvalhais, Deep learning and process understanding for data-driven Earth system science, Nature, 566, (2019) 195-2004.


\bibitem{Ortiz12}
E. G. Ortiz-Garc\'ia, S. Salcedo-Sanz, C. Casanova-Mateo, A. Paniagua-Tineo and A. Portilla-Figueras, “Accurate local very short-term temperature prediction based on synoptic situation support vector regression banks”, Atmos. Re., 107 (2012) 1-8.  

\bibitem{Moosavi15}
V. Moosavi, A. Talebi, M. H. Mokhtari, S. R. Shamsi, Y. Niazi, A wavelet-artificial intelligence fusion approach (WAIFA) for blending Landsat and MODIS surface temperature, Remote Sens. Env., 169 (2015) 243-254.

\bibitem{Xia19}
H. Xia, Y. Chen, Y. Li, J. Quan, Combining kernel-driven and fusion-based methods to generate daily high-spatial-resolution land surface temperatures, Remote Sens. Env., 224 (2019) 259-274.

\bibitem{Zhang20}
Z. Zhang, X. Pan, T. Jiang, B. Sui, C. Liu, W. Sun, Monthly and quarterly sea surface temperature prediction based on gated recurrent unit neural network. J. Mar. Sci. Eng. 8 (2020) 249.



\bibitem{Park17}
S. Park, J. Im, S. Park, J. Rhee, Drought monitoring using high resolution soil moisture through multi-sensor satellite data fusion over the Korean peninsula, Agricult. Forest Met., 237 (2017) 257-269.

\bibitem{Yao17}
Y. Yao, S. Liang, X. Li, J. Chen, S. Liu, J. Jia et al., Improving global terrestrial evapotranspiration estimation using support vector machine by integrating three process-based algorithms, Agricult. Forest Met., 242 (2017) 55-74.

\bibitem{Alizadeh18}
M. R. Alizadeh, M. R. Nikoo, A fusion-based methodology for meteorological drought estimation using remote sensing data, Remote Sens. Env., 211 (2018) 229-247.

\bibitem{Feng19}
P. Feng, B. Wang, D. L. Liu, Q. Yu, Machine learning-based integration of remotely-sensed drought factors can improve the estimation of agricultural drought in South-Eastern Australia, Agricult. Syst., 173 (2019) 303-316.

\bibitem{Dona15}
C. Do\~na, N. B. Chang, V. Caselles, J. M. S\'anchez, B. W. Vannah, Integrated satellite data fusion and mining for monitoring lake water quality status of the Albufera de Valencia in Spain, J. Env. Manag., 151 (2015) 416-426.

\bibitem{Jiang18}
S. Jiang, Y. Zheng, V. Babovic, Y. Tian, F. Han, A computer vision-based approach to fusing spatiotemporal data for hydrological modeling, J. Hydrol., 567 (2018) 25-40.

\bibitem{Sagan20}
V. Sagan, K. T. Peterson, M. Maimaitijiang, P. Sidike, J. Sloan, B. A. Greeling, et al., Monitoring inland water quality using remote sensing: potential and limitations of spectral indices, bio-optical simulations, machine learning, and cloud computing. Earth-Sci. Rev., in press (2020) 103187.



\bibitem{Liu18}
S. Liu, M. Li, Z. Zhang, B. Xiao, X. Cao, Multimodal ground-based cloud classification using joint fusion convolutional neural network, Remote Sens., 10(6) (2018) 822.

\bibitem{Li19}
Z. Li, H. Shen, Q. Cheng, Y. Liu, S. You, Z. He, Deep learning based cloud detection for medium and high resolution remote sensing images of different sensors, ISPRS J. Photogram. Remote Sens., 150 (2019) 197-212.


\bibitem{Wang16}
X. Y. Wang, Y. G. Guo, J. He, L. T. Du, Fusion of HJ1B and ALOS PALSAR data for land cover classification using machine learning methods, Int. J. App. Earth Obs. Geoinf., 52 (2016) 192-203.

\bibitem{Puttina17}
S. Puttinaovarat, P. Horkaew, Urban areas extraction from multi sensor data based on machine learning and data fusion, Patt. Rec. Image Anal., 27 (2017) 326-337.

\bibitem{Guy18}
F. K. Guy, T. Akram, B. Laurent, S. R. Naqvi, M. M. Alex, N. Muhammad, A deep heterogeneous feature fusion approach for automatic land-use classification, Inf. Sci., 467 (2018) 199-218.

\bibitem{Lu19}
Y. Lu, P. Wu, X. Ma, X. Li, Detection and prediction of land use/land cover change using spatiotemporal data fusion and the Cellular Automata--Markov model, Env. Monit. Assess., (2019) in press.

\bibitem{Rasaei19}
Z. Rasaei, P. Bogaert, Spatial filtering and Bayesian data fusion for mapping soil properties: A case study combining legacy and remotely sensed data in Iran, Geoderma, 344 (2019) 50-62.

\bibitem{P197}
N. Yokoya, P. Ghamisi, J. Xia, S. Sukhanov, R. Heremans, I. Tankoyeu, et al., Open Data for Global Multimodal Land Use Classification: Outcome of the 2017 IEEE GRSS Data Fusion Contest, IEEE J. Selected Topics App. Earth Obs. Remote Sens., 11(5) (2018) 1363-1377.

\bibitem{Shaharum20}
N. Shaharum, H. Shafri, W. Ghani, S. Samsatli, M. Al-Habshi, B. Yusuf, Oil palm mapping over Peninsular Malaysia using Google Earth Engine and machine learning algorithms, Remote Sens. App.: Society Env., 17 (2020) 100287.


\bibitem{Alajlan12}
N. Alajlan, Y. Bazi, F. Melgani, R. R. Yager, Fusion of supervised and unsupervised learning for improved classification of hyperspectral images, Inf. Sci., 217 (2012) 39-55.

\bibitem{Ghamisi17}
P. Ghamisi, B. H\"ofle, X. X. Zhu, Hyperspectral and LiDAR Data Fusion Using Extinction Profiles and Deep Convolutional Neural Network, IEEE J. Selected Topics App. Earth Obs. Remote Sens., 10 (2017) 3011-3024.

\bibitem{Khoda15}
M. Khodadadzadeh, J. Li, S. Prasad, A. Plaza, Fusion of hyperspectral and LiDAR remote sensing data using multiple feature learning, IEEE J. Selected Topics App. Earth Obs. Remote Sens., 8 (2015) 2971-2983.

\bibitem{Xu18}
X. Xu, W. Li, Q. Ran, Q. Du, L. Gao, B. Zhang, Multisource Remote Sensing Data Classification Based on Convolutional Neural Network, IEEE Trans. Geosci. Remote Sens., 56 (2018) 937-949.

\bibitem{Liang18}
M. Liang, L. Jiao, S. Yang, F. Liu, B. Hou, H. Chen, Deep Multiscale Spectral-Spatial Feature Fusion for Hyperspectral Images Classification, IEEE J. Selected Topics App. Earth Obs. Remote Sens., 11 (2018) 2911-2924.


\bibitem{Mellit08}
A. Mellit, S. A. Kalogirou, Artificial intelligence techniques for photovoltaic applications: a review, Prog. Energy Comb. Sci., 34(5) (2008) 574-632.

\bibitem{Salcedo14}
S. Salcedo-Sanz, C. Casanova-Mateo, J. Mu\~noz-Mar\'i, G. Camps-Valls, Efficient prediction of daily global solar irradiation using temporal Gaussian processes, IEEE Geosci. Remote Sens. Let., 11(11) (2014) 1136-1140.

\bibitem{Mazorra16}
L. Mazorra-Aguiar, B.Pereira, P.Lauret, F.D\'iaz, M.David, Combining solar irradiance measurements, satellite-derived data and a numerical weather prediction model to improve intra-day solar forecasting, Ren. Energy, 97 (2016) 599-610.

\bibitem{Urraca20}
R. Urraca, T. Huld, A. Gracia-Amillo, F. J. Mart\'inez-de-Pison, F. Kaspar, A. Sanz-Garc\'ia, Evaluation of global horizontal irradiance estimates from ERA5 and COSMO-REA6 reanalyses using ground and satellite-based data, Solar Energy, 164 (2018) 339-354.

\bibitem{Babar20}
B. Babar, L. T. Luppino, T. Bostr\"om, S. N. Anfinsen, Random forest regression for improved mapping of solar irradiance at high latitudes, Solar Energy, 198 (2020) 81-92.


\bibitem{thenkabail2018advanced}
P. S. Thenkabail, J. G. Lyon, A. Huete, Advanced applications in remote sensing of agricultural crops and natural vegetation, CRC press, 2018.

\bibitem{kandasamy2013comparison}
S. Kandasamy, F. Baret, A. Verger, P. Neveux, M. Weiss, A comparison of methods for smoothing and gap filling time series of remote sensing observations--application to modis lai products, Biogeosci., 10 (2013) 4055-4071.

\bibitem{gao2006blending}
F. Gao, J. Masek, M. Schwaller, F. Hall, On the blending of the landsat and modis surface reflectance: Predicting daily landsat surface reflectance, IEEE Trans. Geosci. Remote sens., 44 (2006) 2207-2218.

 \bibitem{sedano2014Kalman}
F. Sedano, P. Kempeneers, G.Hurtt, A kalman filter-based method to generate continuous time series of medium-resolution ndvi images, Remote Sens., 6 (2014) 12381-12408.

\bibitem{Kalman1960new}
R. E. Kalman, A new approach to linear filtering and prediction problems, J. basic Eng., 82 (1960) 35-45.



\bibitem{P7}
L. Estes, D. McRitchie, J. Choi, S. Debats, T. Evans, W. Guthe, et al., A platform for crowdsourcing the creation of representative, accurate landcover maps,
Env. Model. Soft., 80 (2016) 41-53.

\bibitem{P9}
J. Li, J. A. Benediktsson, B. Zhang, T. Yang, A. Plaza, Spatial technology and social media in remote sensing: A survey, Proc. IEEE, 105(10) (2017) 1855–1864.

\bibitem{P192}
M. Chi, Z. Sun, Y. Qin, J. Shen, J. A. Benediktsson, A novel methodology to label urban remote sensing images based on location- IEEE GRSM DRAFT 2018 26 based social media photos, Proc. IEEE, 105(10) (2017) 1926-1936.

\bibitem{P193}
J. M. Shapiro, Smart cities: Quality of life, productivity, and the growth effects of human capital, Rev. Econ. Stat., 88(2) (2006) 324–335.

\bibitem{P194}
P. Gamba, Human settlements: A global challenge for EO data processing and interpretation, Proc. IEEE, 101(3) (2013) 570–581.

\bibitem{P195}
M. T. Eismann, A. D. Stocker, N. M. Nasrabadi, Automated hyperspectral cueing for civilian search and rescue, Proc. IEEE, 97(6) (2009) 1031–1055.

\bibitem{P196}
S. B. Serpico, S. Dellepiane, G. Boni, G. Moser, E. Angiati, R. Rudari, Information extraction from remote sensing images for flood monitoring and damage evaluation, Proc. IEEE, 100(10) (2012) 2946–2970.



\bibitem{Fisher98}
R. Fisher, J. Fulcher, Improving the inversion of ionograms by combining neural network and data fusion techniques, Neural Comput. App., 7 (1998) 3-16.

\bibitem{Carro17}
L. Carro-Calvo, C. Casanova-Mateo, J. Sanz-Justo, P. Salvador, S. Salcedo-Sanz, Efficient prediction of total column ozone based on Support Vector Regression algorithms, numerical models and Suomi-satellite data”, Atm\'osfera Journal, 30 (2017) 1-10.

\bibitem{Du19}
Y. Du, W. Song, Q. He, D. Huang, A. Liotta, C. Su, Deep learning with multi-scale feature fusion in remote sensing for automatic oceanic eddy detection, Inf. Fus., 49 (2019) 89-99.

\bibitem{Kattenborn15}
T. Kattenborn, J. Maack, F. Fa$\beta$nacht, F. En$\beta$le, J. Ermert, B. Koch, Mapping forest biomass from space -- Fusion of hyperspectral EO1-hyperion data and Tandem-X and WorldView-2 canopy height models, International Journal of App. Earth Obs. Geoinform., 35(B) (2015) 359-367.

\bibitem{Effrosy18}
D. Effrosynidis, A. Arampatzis, G. Sylaios, Seagrass detection in the mediterranean: A supervised learning approach, Ecol. Inform., 48 (2018) 158-170.

\bibitem{Li18b}
S. Li, Y. Yao, J. Hu, G. Liu, X. Yao, J. Hu, An ensemble stacked convolutional neural network model for environmental event sound recognition, Appl. Sci., 8 (2018) 1152.




\bibitem{Danson95}
F. M. Danson, S. E. Plummer, Advances in Environmental Remote Sensing, John Wiley \& Sons, New York, 1995.

\bibitem{Richards99}
J. A. Richards, X. Jia, Remote Sensing Digital Image Analysis. An Introduction, Springer-Verlag, Berlin, 1999.

\bibitem{Ustin04}
S. Ustin, Remote Sensing for Natural Resource Management and Environmental Monitoring, Manual of Remote Sensing, 4, John Wiley \& Sons, New York, 2004.

\bibitem{Liang04}
S. Liang, Quantitative Remote Sensing of Land Surfaces,John Wiley \& Sons, New York, 2004.

\bibitem{lillesand08}
T. M. Lillesand, R. W. Kiefer, J. Chipman, Remote Sensing and Image Interpretation, John Wiley \& Sons, New York, 2008.

\bibitem{Mott07}
H. Mott, Remote Sensing with Polarimetric Radar, John Wiley \& Sons, New York, 2007.

\bibitem{Wang08}
B. C. Wang, Digital Signal Processing Techniques and Applications in Radar Image	Processing, John Wiley \& Sons, New York, 2008.

\bibitem{shaw02}
G. Shaw and D. Manolakis, Signal Processing for Hyperspectral Image Exploitation, Signal Proc. Mag., 50 (2002) 12-16.



\bibitem{Klein02}
A. M. Klein-Tank, et al. Daily dataset of 20th-century surface air temperature and precipitation series for the European climate assessment. Int. J. Climatol., 22(12) (2002) 1441-1453.

\bibitem{ECAD}
https://www.ecad.eu/

\bibitem{CRU}
https://crudata.uea.ac.uk/cru/data/temperature/

\bibitem{GPCC}
https://opendata.dwd.de/climate$\_$environment/GPCC/html/download$\_$gate.html

\bibitem{GPCP}
https://www.esrl.noaa.gov/psd/data/gridded/data.gpcp.html

\bibitem{GHCN}
https://www.esrl.noaa.gov/psd/data/gridded/data.ghcngridded.html

\bibitem{GISS}
https://www.esrl.noaa.gov/psd/data/gridded/data.gisstemp.html

\bibitem{ESRL}
https://www.esrl.noaa.gov/psd/cgi-bin/db\_search/SearchMenus.pl



\bibitem{Baldocchi2008}
D. Baldocchi, Breathing of the terrestrial biosphere: lessons learned from a global network of carbon dioxide flux measurement systems, Australian J. Botany, 56 (2008) 1-26.

\bibitem{Tramontana16bg}
G. Tramontana, M. Jung,C. R. Schwalm, K. Ichii, G. Camps-Valls, B. R\`aduly, et al., Predicting carbon dioxide and energy fluxes across global FLUXNET sites with regression algorithms, Biogeosci., 13 (2016) 4291–4313.

\bibitem{Jung18fluxcom}
M. Jung, S. Koirala, U. Wever, K. Ichii, F. Gans, G. Camps-Valls, et al. The FLUXCOM ensemble of global land-atmosphere energy fluxes, Sci. Data, 6(74) 2019.

\bibitem{Mahecha19esdc}
M. Mahecha, F. Gans, G. Brandt, R. Christiansen, S. Cornell, N. Fomferra, et al., Earth system data cubes unravel global multivariate dynamics, Earth Syst. Dyn., Submitted, in open discussion and review, 2019.



\bibitem{RUC}
https://ruc.noaa.gov/raobs/

\bibitem{UWYO}
http://weather.uwyo.edu/upperair/sounding.html


\bibitem{NDBC}
https://www.ndbc.noaa.gov/

\bibitem{PEIOMDM}
https://www.seadatanet.org/

Australian Ocean Data network

\bibitem{OpenAccAus}
https://portal.aodn.org.au/search




\bibitem{Kanamitsu89}
M. Kanamitsu, Description of the NMC global data assimilation and forecast system. Weath. Forecast., 4 (1989) 335-342.

\bibitem{Kanamitsu91}
M. Kanamitsu, J.C. Alpert, K.A. Campana, P.M. Caplan, D.G. Deaven, M. Iredell, et al. Recent changes implemented into the global forecast system at NMC. Weath. Forecast., 6 (1991) 425-435.

\bibitem{Cote98}
J. C\^ot\'e, J. G. Desmarais, S. Gravel, A. M\'ethot, A. Patoine, M. Roch, and A. Staniforth, The operational CMC-MRB Global Environmental Multiscale (GEM) model: Part II - Results, Month. Weath. Rev., 126 (1998) 1397-1418.

\bibitem{Cote98b}
J. C\^ot\'e, S. Gravel, A. M\'ethot, A. Patoine, M. Roch, A. Staniforth, The operational CMC-MRB Global Environmental Multiscale (GEM) model: Part I - Design considerations and formulation, Month. Weath. Rev., 126 (1998) 1373-1395.

\bibitem{Yeh02}
K. S. Yeh, J. C\^ot\'e, S. Gravel, A. M\'ethot, A. Patoine, M. Roch, A. Staniforth, The CMC-MRB Global Environmental Multiscale (GEM) Model. Part III: Nonhydrostatic Formulation. Month. Weath. Rev., 130(2) (2002) 339-356.

\bibitem{Hogan14}
T. F. Hogan, M. Liu, J.A. Ridout, M.S. Peng, T.R. Whitcomb, B.C. Ruston, The Navy Global Environmental Model, Oceanography, 27 (2014) 116-125.

\bibitem{ECMWF}
https://www.ecmwf.int/en/forecasts/documentation-and-support

\bibitem{Walters17}
D. Walters, I. Boutle, M. Brooks, T. Melvin, R. Stratton, S. Vosper, The Met Office Unified Model Global Atmosphere 6.0/6.1 and JULES Global Land 6.0/6.1 configurations, Geosci. Model Dev., 10 (2017) 1487-1520.

\bibitem{Deque94}
M. D\'equ\'e, C. Dreveton, A. Braun, D. Cariolle, The ARPEGE/IFS atmosphere model: a contribution to the French community climate modelling, Clim. Dyn.  10 (1994) 249-266.

\bibitem{Buizza05}
R. Buizza, P. L. Houtekamer, Z. Toth, G. Pellerin, M. Wei, Y. Zhu, A comparison of the ECMWF, MSC and NCEP global ensemble prediction systems, Month. Weath. Rev., 133 (2005) 1076-1097.

\bibitem{Perez13}
R. P\'erez, E. Lorenz, S. Pelland, M. Beauharnois, G. Van Knowe, K. Hemker Jr., et al., Comparison of numerical weather prediction solar irradiance forecasts in the US, Canada and Europe, Sol. Energy, 94 (2013) 305-326.

\bibitem{Dueben18}
P. D. Dueben, P. Bauer, Challenges and design choices for global weather and climate models based on Machine Learning, Geosci. Model Dev., 11 (2018) 3999-4009.



\bibitem{Kalnay96}
E. Kalnay et al. The NCEP/NCAR 40-Year Reanalysis project. Bull. Amer. Meteor. Soc., 77 (1996) 437-471.

\bibitem{Kanamitsu02}
M. Kanamitsu, et al. NCEP-DOE AMIP-II Reanalysis (R-2),  Bull. Amer. Meteor. Soc., 83 (2002) 1631-1643.

\bibitem{Saha10}
S. Saha et. al. The NCEP Climate Forecast System Reanalysis, Bull. Amer. Meteor. Soc., 91 (2010) 1015-1057.

\bibitem{ERA_Interim}
D. P. Dee, S. M Uppala, A. J. Simmons, P. Berrisford, P. Poli, S. Kobayashi, et al. The ERA-Interim reanalysis: Configuration and performance of the data assimilation system, Q. J. Royal Met. Soc., 137 (2011) 553-597.

\bibitem{ERA5}
https://www.ecmwf.int/en/forecasts/datasets/reanalysis-datasets/era5

\bibitem{Rienecker11}
M. M. Rienecker et al., MERRA: NASA's Modern-Era Retrospective Analysis for Research and Applications. J. Climate, 24 (2011) 3624-3648.

\bibitem{JRA25}
K. Onogi et al., The JRA-25 Reanalysis. J. Met. Soc. Jap. 85(3) (2007) 369-432.

\bibitem{Mesinger06}
F. Mesinger, et al. North American Regional Reanalysis. Bull. American Met. Soc., 87 (2006) 343-360.

\bibitem{Bollmeyer15}
C. Bollmeyer, et al. Towards a high-resolution regional reanalysis for the European CORDEX domain, Q. J. Royal Met. Soc., 141 (2015) 1-15.

\bibitem{Euro4M}
http://www.euro4m.eu/

\bibitem{Hodges11}
Hodges, K. I., R. W. Lee, L. Bengtsson, A comparison of extratropical cyclones in recent reanalyses ERA-Interim,
NASA MERRA, NCEP CFSR, and JRA-25. J. Climate, 24 (2011) 4888-4906.

\bibitem{Bao12}
X. Bao, F. Zhang, Evaluation of NCEP-CFSR, NCEP-NCAR, ERA-Interim, and ERA-40 Reanalysis
Datasets against Independent Sounding Observations over the Tibetan Plateau, J. Climate, 26 (2012) 206-214.

\bibitem{Chaudhuri14}
A. H. Chaudhuri, R. M. Ponte, T. Nguyen, A Comparison of Atmospheric Reanalysis Products for the Arctic Ocean
and Implications for Uncertainties in Air-Sea Fluxes, J. Climate, 27 (2014) 5411-5421.





























\bibitem{GCOS}
GCOS secretariat, Guideline for the Generation of Satellite-based Datasets and Products meeting GCOS Requirements,Global Climate Observing System, 2010, http://www.wmo.int/pages/prog/gcos/index.php

\bibitem{Pipia19}
L. Pipia, J. Muñoz-Mar\'i, E. Amin, S. Belda, G. Camps-Valls, J. Verrelst, Fusing optical and SAR time series for LAI gap filling with multioutput Gaussian processes, Remote Sens. Env., 235 (2019) 111452.

\bibitem{Alvarez11}
M. A. \'Alvarez, L. Rosasco, N. D. Lawrence, Kernels for vector-valued functions: a review, arXiv:1106.6251 [cs, math, stat]  (2011).

\bibitem{Dorigo2017}
W. Dorigo, W. Wagner, C. Albergel, F. Albrecht, G. Balsamo, L. Brocca, et al., ESA CCI soil moisture for improved earth system understanding: State-of-the art
 and future directions, Remote Sens. Env. 203 (2017) 185-215.

\bibitem{Sanchez12}
N. S\'anchez, J. Mart\'inez-Fern\'andez, A. Scaini, C. P\'erez-Guti\'errez, Validation of the SMOS L2 soil moisture data in the {REMEDHUS} network (Spain), IEEE Trans.s Geosci. Remote Sens., 50(5) (2012) 1602-1611.

\bibitem{dahra}
T. Torbern, F. Rasmus, G. Idrissa, R. M. Olander, H. Silvia, M. Cheikh, et al., Ecosystem properties of semiarid savanna grassland in west Africa and its
  relationship with environmental variability,
Glob. Change Biol., 21(1) (2014) 250-264.



\bibitem{gevaert2015comparison}
C. M. Gevaert, C.M., F. J. Garc{\'\i}a-Haro, A comparison of starfm and an unmixing-based algorithm for landsat and modis data fusion, Remote Sens. Env., 156 (2015) 34-44.

\bibitem{gorelick2017google}
N. Gorelick, M. Hancher, M., M. Dixon, S. Ilyushchenko, D. Thau, R. Moore, Google earth engine: Planetary-scale geospatial analysis for everyone, Remote Sens. Env., 202 (2017) 18-27.

\bibitem{dee1998data}
D. P. Dee, A. M. Da~Silva, Data assimilation in the presence of forecast bias, Q. J. Royal Met. Soc., 124 (1998) 269-295.

\bibitem{AP1}
H, Ji, X. Luo, 3D scene reconstruction of landslide topography based on data fusion between laser point cloud and UAV image, Environ. Earth Sci., 78 (2019) 534.

\bibitem{AP2}
A. C. Oliveira, L. C. Botega, J. F. Saran, J. N. Silva, J. O. Melo, M. F. Tavares, et al., Crowdsourcing, data and information fusion and situation awareness for emergency Management of forest fires: The project DF100Fogo (FDWithoutFire), Comput. Environ. Urban Syst., 77 (2019) 101172.

\bibitem{AP3}
B. Tang, J. Tang, Y, Liu, F. Zeng, Comprehensive Evaluation and Application of GIS Insulation Condition Part 1: Selection and Optimization of Insulation Condition Comprehensive Evaluation Index Based on Multi-Source Information Fusion, IEEE Access, 7 (2019) 88254-88263.

\bibitem{P20}
B. Choubin, M. Abdolshahnejad, E. Moradi, X. Querol, A. Mosavi, S. Shamshirband, et al., Spatial hazard assessment of the PM10 using machine learning models in Barcelona, Spain, Sci. Total Env., 701 (2020) 134474.

\bibitem{P21}
B. Choubin, A. Mosavi, E. H. Alamdarloo, F. S. Hosseini, S. Shamshirband, K. Dashtekian, et al., Earth fissure hazard prediction using machine learning models, Env. Res, 179 (2019) 108770.


\bibitem{AP4}
C. Shi, X. Wang, M. Zhang, X. Liang, L. Niu, H. Han, A comprehensive and automated fusion method: The enhanced flexible spatiotemporal data fusion model for monitoring dynamic changes of land surface, Appl. Sci., 9(18) (2019) 3693.

\bibitem{AP5}
X. Shan, H. Yin, X. Liu, Z. Wang, C. Qu, G. Zhang, et al. High-rate real-time GNSS seismology and early warning of earthquakes’, Acta Geophys. Sin., 2(8) (2019) 3043-3052.

\bibitem{AP6}
Y. Yang, J. Yang, C. Xu, C. Xu, C. Song, Local-scale landslide susceptibility mapping using the B-GeoSVC model, Landslides, 16(7) (2019) 1301-1312.

\bibitem{AP7}
P. Feng, B. Wang, D. L. Liu, Q. Yu, Machine learning-based integration of remotely-sensed drought factors can improve the estimation of agricultural drought in South-Eastern Australia, Agric. Syst., 173 (2019) 303-316.

\bibitem{AP8}
Y. Zou, S. M. O’Neill, N. Larkin, E. C. Alvarado, R. Solomon, C. Mass, et al., Machine learning-based integration of high-resolution wildfire smoke simulations and observations for regional health impact assessment, Int. J. Environ. Res. Public Health, 16(12) (2019) 2137.

\bibitem{AP9}
K. R. Knipper, W. P. Kustas, M. C. Anderson, J. G. Alfieri, J. H. Prueger, C. R. Hain, Evapotranspiration estimates derived using thermal-based satellite remote sensing and data fusion for irrigation management in California vineyards, Irrig. Sci., 37(3) (2019) 431-449.

\bibitem{AP10}
L. Guerriero, A. Cusano, G. Ruzza, P. Revellino, F. M. Guadagno, Flood hazard mapping in convex floodplain: Multiple probability models fusion, bank threshold and levees effect spatialization, Ital. J. Eng. Geol. Environ., 2019 (2019) 47-52.

\bibitem{AP11}
C. Lee, I. Tien, Probabilistic Framework for Integrating Multiple Data Sources to Estimate Disaster and Failure Events and Increase Situational Awareness, ASCE-ASME J. Risk Uncertain. Eng. Syst. Part A. Civ. Eng., 4(4) (2018) 04018042.

\bibitem{AP12}
M. Azmi, C. R\"udiger, Validating the data fusion-based drought index across Queensland, Australia, and investigating interdependencies with remote drivers, Int. J. Climatol., 38(11) (2018) 4102-4115.

\bibitem{AP13}
M. R. Alizadeh, M.R. Nikoo, A fusion-based methodology for meteorological drought estimation using remote sensing data, Remote Sens. Environ., 211 (2018) 229-247.

\bibitem{AP14}
N. J. Pastick, B. K. Wylie, Z. Wu, Spatiotemporal analysis of Landsat-8 and Sentinel-2 data to support monitoring of dryland ecosystems, Remote Sens. 10(5) (2018) 791.

\bibitem{AP15}
H. Li, Q. Fan, Q. Data acquisition and mining of landslide risk sensing data based on unmanned aerial vehicle, Tech. Bull., 55(10) (2017) 193-199.

\bibitem{AP16}
L. Zhuo, D. Han, Multi-source hydrological soil moisture state estimation using data fusion optimisation, Hydrol. Earth Syst. Sci., 21(7) (2017) 3267-3285.

\bibitem{AP17}
J. F. Rosser, D. G. Leibovici, M. J. Jackson, Rapid flood inundation mapping using social media, remote sensing and topographic data, Nat. Hazards, 87(1) (2017) 103-120.

\bibitem{AP18}
C. S. Renschler, Z. Wang, Multi-source data fusion and modeling to assess and communicate complex flood dynamics to support decision-making for downstream areas of dams: The 2011 hurricane Irene and schoharie creek floods, NY, Int. J. Appl. Earth Obs. Geoinform., 62 (2017) 157-173.

\bibitem{AP19}
F. Hillen, Geo-information fusion for time-critical geo-applications during major events, Gis. Sci. - Die Z. Geoinformatik, 1 (2017) 1-9.

\bibitem{AP20}
A. D'Addabbo, A. Refice, F. P. Lovergine, G. Pasquariello, DAFNE: A Matlab toolbox for Bayesian multi-source remote sensing and ancillary data fusion, with application to flood mapping, Comput. Geosci., 112 (2018) 64-75.

\bibitem{AP21}
Y. Guo, X. Jia, D. Paull, J. A. Benediktsson, Nomination-favoured opinion pool for optical-SAR-synergistic rice mapping in face of weakened flooding signals, ISPRS J. Photogramm. Remote Sens., 155 (2019) 187-205.

\bibitem{AP25}
H. Shafizadeh-Moghadam, R. Valavi, H. Shahabi, K. Chapi, A. Shirzadi, Novel forecasting approaches using combination of machine learning and statistical models for flood susceptibility mapping, J. Environ. Manag., 217 (2018) 1-11.

\bibitem{AP26}
H. Darabi, B. Choubin, O. Rahmati, A. Torabi-Haghighi, B. Pradhan, B. Kl{\o}ve, Urban flood risk mapping using the GARP and QUEST models: A comparative study of machine learning techniques, J. Hydrol., 569 (2019) 142-154.

\bibitem{AP27}
S. A. Woznicki, J. Baynes, S. Panlasigui, M. Mehaffey, A. Neale,  Development of a spatially complete floodplain map of the conterminous United States using random forest, Sci. Total Environ. 647 (2019) 942-953.

\bibitem{AP28}
M. Clement, C. Kilsby, P. Moore, Multi‐temporal synthetic aperture radar flood mapping using change detection, J. Flood Risk Manag., 11(2) (2018) 152-168.



\bibitem{buchanan2009delivering}
G. Buchanan, A. Nelson, P. Mayaux, A. Hartley, P. F. Donald, Delivering a global, terrestrial, biodiversity observation system through remote sensing, Conserv. Biol., 23(2) (2009) 499-502.

\bibitem{giri2011status}
C. Giri, E. Ochieng, L. Tieszen, Z. Zhu, A. Singh, T. Loveland, et al., Status and distribution of mangrove forests of the world using earth observation satellite data, Global Ecol. Biogeograph., 20(1) (2011) 154-159.

\bibitem{skidmore2003environmental}
A. Skidmore, Environmental modelling with GIS and remote sensing, CRC Press, 2003.

\bibitem{turner2003methodological}
M. D. Turner, Methodological reflections on the use of remote sensing and geographic information science in human ecological research, Human ecol., 31(2) (2003) 255-279.

\bibitem{huang2018urban}
B. Huang, B. Zhao, Y. Yimeng, Urban land-use mapping using a deep convolutional neural network with high spatial resolution multispectral remote sensing imagery, Remote Sensing Environ., 214 (2018) 73-86.

\bibitem{zheng2018survey}
X. Zheng, J. Han, A. Sun, A survey of location prediction on twitter, IEEE Trans. Knowledge Data Eng., 30(9) (2018) 1652-1671.

\bibitem{Leichter18}
A. Leichter, D. Wittich, F. Rottensteiner, M. Werner, M. Sester, Improved classification of satellite imagery using spatial feature maps extracted from social media, ISPRS - International Archives of the Photogrammetry, Remote Sensing Spatial Inform. Sci., XLII-4 (2018) 335-342.

\bibitem{li2013spatial}
L. Li, M. F. Goodchild, B. Xu, Spatial, temporal, and socioeconomic patterns in the use of Twitter and Flickr, Cartograph. Geograph. Inf. Sci., 40(2) (2013) 61-77.

\bibitem{lott2012survey}
B. Lott, Survey of keyword extraction techniques, UNM Education, 50 (2012) 1-11.

\bibitem{chen2019experimental}
Y. Chen, H. Zhang, R. Liu, Z. Ye, J. Lin, Experimental explorations on short text topic mining between LDA and NMF based Schemes, Knowledge-Based Syst., 163 (2019) 1-13.

\bibitem{manna2019effectiveness}
S. Manna, H. Naki, Effectiveness of Word Embeddings on Classifiers: A Case Study with Tweets, Proc. of the IEEE 13th International Conference on Semantic Computing (ICSC), (2019) 158-161.

\bibitem{tang2014learning}
D. tang, F. Wei, N. Yang, M. Zhou, T. Liu, B. Qin, Learning sentiment-specific word embedding for twitter sentiment classification, Proc. of the 52nd Annual Meeting of the Association for Computational Linguistics, (2014) 1555-1565.

\bibitem{Boja16}
P. Bojanowski, E. Grave, A. Joulin, T. Mikolov,  Trans. Assoc. Comput. Linguistics, 5 (2017) 135–146.

\bibitem{wang2015unified}
P. Wang, Y. Qian, F. K. Soong, L. He, H. Zhao, A unified tagging solution: Bidirectional lstm recurrent neural network with word embedding, arXiv preprint arXiv:1511.00215, 2015.

\bibitem{igarssbtclassification}
M H\"aberle, M. Werner, X. Xiaoxiang, Building Type Classification From Social Media Texts via Geo-Spatial Text Mining, Proc. of the International Geoscience and Remote Sensing Symposium (IGARSS'19), 2019.

\bibitem{iosifidis2019sentiment}
V. Iosifidis, E. Ntoutsi, Sentiment analysis on big sparse data streams with limited labels, Knowl. Inf. Syst., in press, 2019.

\bibitem{igarssmutualinformation}
E. J. Eike, M. Werner and X. Zhu, Mutual Information Analysis of Social Media Images and Building Functions,  Proc. of the International Geoscience and Remote Sensing Symposium, 2019.

\bibitem{dlr126371}
E. J. Eike, M. Werner and X. Zhu, Building Instance Classification using Social Media Images, Proc. of the Joint Urban Remote Sensing Event (JURSE), 2019.

\bibitem{groundaerialfusion}
E. J. Hoffmann, Y. Wang, M. Werner, J. Kang, X. X. Zhu, Model Fusion for Building Type Classification from Aerial and Street View Images, Remote Sens., 11(11) (2019) 1259.

\bibitem{cao2018integrating}
R. Cao, J. Zhu, W. Tu, Q. Li, J. Cao, B. Liu, Qet al., Integrating Aerial and Street View Images for Urban Land Use Classification, Remote Sens., 10(10) (2018) 1553.


















\end{thebibliography}
